%% file: main.tex
\documentclass{article}

\usepackage{iclr2023_conference}
\usepackage{times}
\usepackage{xcolor}
\usepackage{xspace}
\usepackage[pdftex]{graphicx}
\usepackage{amsmath,amssymb,amsfonts,dsfont,pifont,bm,bbm,mathrsfs,mathtools,nicefrac}
\usepackage{algorithm,algpseudocode,listings}
\usepackage{booktabs,multirow,adjustbox,diagbox,threeparttable}
\definecolor{citeblue}{RGB}{48,111,186}
\usepackage[pagebackref=false,breaklinks=true,colorlinks=true,citecolor=citeblue,bookmarks=false]{hyperref}
\usepackage{cleveref}  % Should be loaded after 'hyperref', and works perfectly with 'subfigure'.
\usepackage{capt-of}
\usepackage{wrapfig}
\crefname{section}{Sec.}{Secs.}
\Crefname{section}{Section}{Sections}
\crefname{table}{Tab.}{Tabs.}
\Crefname{table}{Table}{Tables}
\crefname{figure}{Fig.}{Figs.}
\Crefname{figure}{Figure}{Figures}
\crefname{equation}{Eq.}{Eqs.}
\Crefname{equation}{Equation}{Equations}
\input{math_commands.tex}

\definecolor{revision_color}{rgb}{0, 0, 0}
\newcommand{\rebuttal}[1]{\textcolor{black}{#1}}

\newcommand{\tocite}[1]{\textcolor{red}{[TO CITE]}}

\title{Towards Smooth Video Composition}

\author{Qihang Zhang$^1$\quad Ceyuan Yang$^2$\quad Yujun Shen$^3$\quad Yinghao Xu$^1$\quad Bolei Zhou$^4$ \\ $^1$The Chinese University of Hong Kong,\quad $^2$Shanghai AI Laboratory, \quad$^3$Ant Group,\\$^4$University of California, Los Angeles}

\iclrfinalcopy % Uncomment for camera-ready version, but NOT for submission.

\begin{document}

\maketitle

\begin{center}
    \vspace{-5pt}
	\includegraphics[width=1.0\textwidth]{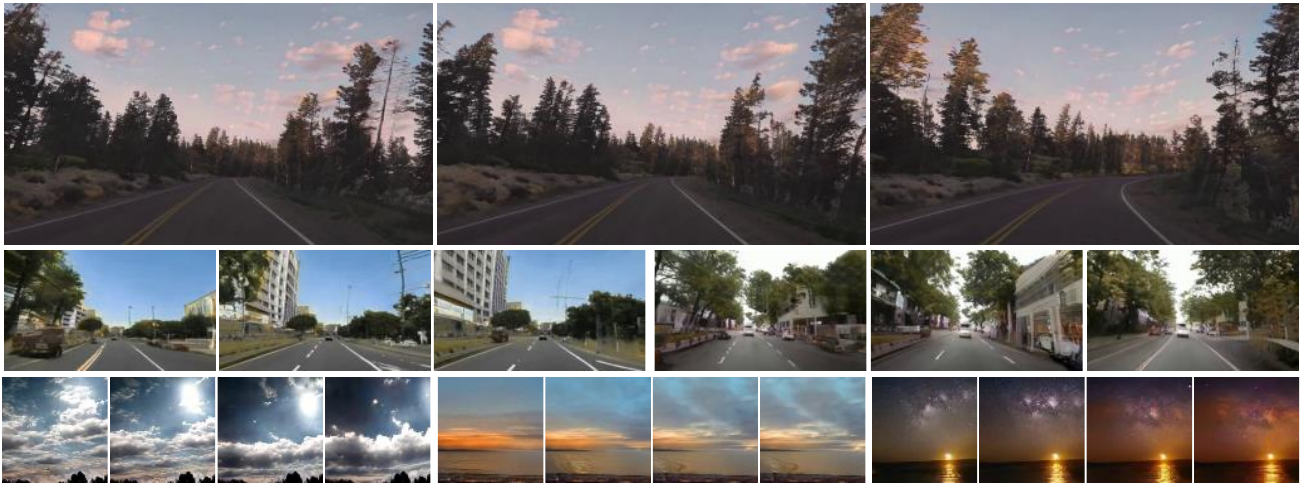}
	\vspace{-20pt}
	\captionof{figure}{
        Generated video frames from the proposed model trained on Countryside, YouTube Driving~\citep{zhang2022learning}, and SkyTimelapse~\citep{xiong2018learning} (from top to bottom).
        %
        % Demo video is in the appendix.
    }
    \label{figure:gallery}
    \vspace{10pt}
\end{center}

\input{sections/0.abs.tex}
\input{sections/1.intro.tex}
\input{sections/2.method.tex}
\input{sections/3.exp.tex}
\input{sections/4.related_work.tex}
\input{sections/5.conclusion.tex}
\input{sections/6.ref.tex}
\input{sections/7.appendix.tex}

\end{document}

%% file: math_commands.tex
\usepackage{amsmath,amsfonts,bm}

\def\eqref#1{equation~\ref{#1}}
\def\1{\bm{1}}

\DeclareMathAlphabet{\mathsfit}{\encodingdefault}{\sfdefault}{m}{sl}
\SetMathAlphabet{\mathsfit}{bold}{\encodingdefault}{\sfdefault}{bx}{n}

%% file: sections/0.abs.tex
\begin{abstract}

Video generation requires synthesizing consistent and persistent frames with dynamic content over time.
This work investigates modeling the temporal relations for composing video with arbitrary length, from a few frames to even infinite, using generative adversarial networks (GANs).
First, towards composing adjacent frames, we show that the alias-free operation for single image generation, together with adequately pre-learned knowledge, brings a smooth frame transition without compromising the per-frame quality.
Second, by incorporating the temporal shift module (TSM), originally designed for video understanding, into the discriminator, we manage to advance the generator in synthesizing more consistent dynamics.
Third, we develop a novel B-Spline based motion representation to ensure temporal smoothness to achieve infinite-length video generation. It can go beyond the frame number used in training.
A low-rank temporal modulation is also proposed to alleviate repeating contents for long video generation.
We evaluate our approach on various datasets and show substantial improvements over video generation baselines. 
Code and models will be publicly available at \url{https://genforce.github.io/StyleSV}.

\end{abstract}

%% file: sections/1.intro.tex
\section{Introduction}\label{sec:intro}

Synthesizing images using generative adversarial network (GAN)~\citep{gan,dcgan, stylegan, stylegan2, karras2021alias, pggan} usually requires composing diverse visual concepts and their spatial arrangement. 
Recent advances in GANs have enabled many appealing applications such as customized editing~\citep{goetschalckx2019ganalyze, shen2020interfacegan, jahanian2019steerability, yang2021semantic} and animation~\citep{qiu2022stylefacev, alaluf2022third}.
However, employing GANs for video generation remains challenging given the additional requirement of modeling the temporal dynamics. 

In fact, a video is not simply a stack of images.
Instead, the content in video frames should have a smooth transition over time, and the video may last arbitrarily long.
Thus, compared to image synthesis, the crux of video synthesis lies in modeling the temporal relations across frames. % and compose them properly. 
We categorize the temporal relations into three folds regarding the time scale.
First, when looking at a transient dynamic, we focus more on the subtle change between neighbor frames and expect decent local motions, such as facial muscle movement and cloud drifting.
When the duration becomes longer, more content may vary across frames.
Under such a case, it is vital to learn a consistent global motion representation.
For example, in a first-view driving video, trees and buildings on the sides of the road should move backward when the car is driving forward.
Finally, in videos that last longer, new objects have to be generated over time. 
Therefore it requires the movements of the newly added objects are consistent and coherent.

This work aims at smooth video composition by modeling multi-scale temporal relations with GANs.
First, we confirm that, same as in image synthesis, the texture sticking problem (\textit{i.e.}, some visual concepts are bound to their coordinates) also exists in video generation, interrupting the smooth flow of content across frames.
To tackle this issue, we deploy the alias-free technique~\citep{karras2021alias} from single image generation and preserve the frame quality via appropriate pre-training.
Then, to make the generator produce realistic motion dynamics, we incorporate a temporal shift module (TSM)~\citep{lin2019tsm} in the discriminator as an inductive bias.
Thus the discriminator captures more information from the temporal dimension for real/fake classification, providing better guidance to the generator.
Furthermore, we observe that the motion representation proposed in previous work~\citep{skorokhodov2022stylegan} suffers from undesired content jittering (see \cref{subsec:gobeyond_limit} for details) for long video generation.
We identify its cause as the first-order discontinuity when interpolating motion embeddings.
To mitigate this issue, we design a novel B-Spline based motion representation that can transit smoothly over time.
A low-rank strategy is further proposed to alleviate the issue that the content across frames may repeat cyclically.
We evaluate the improved models on various video generation benchmarks, including YouTube driving dataset~\citep{zhang2022learning}, SkyTimelapse~\citep{xiong2018learning}, \rebuttal{Taichi-HD~\citep{Siarohin_2019_NeurIPS}} and observe consistent and substantial improvements over existing alternatives.
Given its simplicity and efficiency, our approach brings a simple yet strong baseline for video generation.

%% file: sections/2.method.tex
\section{Method}\label{sec:method}
We introduce several improvements made on the prior art StyleGAN-V~\citep{skorokhodov2022stylegan} to set a new baseline for video generation. 
We first introduce the default configuration (\textbf{Config-A}) of the StyleGAN-V in \cref{subsec:preliminary}.
We then make a comprehensive overhaul of it. Concretely, in \cref{subsec:alias_free_and_pretrain} we confirm that the alias-free technique (\textbf{Config-B}) in single image generation, together with adequately pre-learned knowledge (\textbf{Config-C}), results in a smooth transition of two adjacent frames.
\cref{subsec:tsm} shows that when the temporal information is explicitly modeled into the discriminator through the temporal shift module~\citep{lin2019tsm} (\textbf{Config-D}), the generator produces significantly better dynamic content across frames.
Although prior arts could already generate arbitrarily long videos, the cyclic jittering is observed as a video continues. Thus We propose a B-Spline based motion representation (\textbf{Config-E}) to ensure the continuity, which together with a low-rank temporal modulation (\textbf{Config-F}) produces much more realistic long videos in \cref{subsec:gobeyond_limit}. %

\subsection{Preliminary}\label{subsec:preliminary}

StyleGAN-V~\citep{skorokhodov2022stylegan} introduces continuous motion representations and a holistic discriminator for video generation. 
Specifically, continuous frames $I_t$ could be obtained by feeding continuous $t$ into a generator $G(\cdot)$:
\begin{equation}
    %I_{t} = G(u, v_t) = G(u, f(t)),~t\in [0,+\infty],
    I_{t} = G(u, v_t),
\end{equation}
where $u$ and $v_t$ denote the content code and continuous motion representation. Concretely, content code is sampled from a standard distribution while the motion representation $v_t$ consists of two embeddings: time positional embedding $v_t^{pe}$ and interpolated motion embedding $v_t^{me}$. 

To obtain the time positional embedding $v_t^{pe}$, \rebuttal{we first randomly sample $N_A$ codes} as a set of discrete time anchors $A_i,i \in[0,\cdots,N_{A}-1]$ that share equal interval (256 frames in practice). Convolutional operation with 1D kernel is then applied on anchor $A_i$ for temporal modeling, producing the corresponding features $a_i$ with timestamp $t_i$. For an arbitrary continuous $t$, its corresponding interval is first found with the nearest left and right anchor feature $a_l$ and $a_{l+1}$, at time $t_l$ and $t_{l+1}$, so that $t_l\leq t < t_{l+1}, l\in[0,\cdots,N_{A}-2]$.
For time positional embedding $v_t^{pe}$, a wave function is derived with learned frequency $\beta$, phase $\gamma$ and amplitude $\alpha$ from the left anchor feature $a_l$: 
%
%Then $v_t^{pe}$ is calculated by the wave function with $t$ as input:
%
\begin{equation}
\begin{aligned}
\alpha &= M_{\alpha}(a_l), \beta = M_{\beta}(a_l), \gamma = M_{\gamma}(a_l), \\
v_t^{pe} &= <\alpha\cdot \sin(\beta\cdot t + \gamma), \alpha\cdot \cos(\beta\cdot t + \gamma)>,  \label{eq:pe}
\end{aligned}
\end{equation}
where $M_\alpha, M_\beta, M_\gamma$ are learnable MLPs. 
%
%MLPs mapping from anchor feature $a_i$ to amplitude $\alpha$, frequency $\beta$, and phase $\gamma$ space respectively.
%
The interpolated motion embedding $v_t^{me}$ could be obtained through linear interpolation between anchor feature $a_l$ and $a_{l+1}$ based on $t$:
\begin{equation}
    v_t^{me} = \frac{(t-t_l)a_l+(t_{l+1}-t)a_{l+1}}{t_{l+1}-t_l}.  \label{eq:me}
\end{equation}
The final motion representation $v_t$ is thus the sum of these two terms:
\begin{equation}
    v_t = v_t^{pe} + v_t^{me}.
\end{equation}

Additionally, a sparse sampling strategy is proposed for efficient training. Namely,
$N_t$ discrete time steps $t_i$ and real video frames $I_i$ are sampled respectively. By feeding $N_t$ discrete time steps and one content code into generator, multiple synthesis $I^{\prime}_i = G(u, v_{t_i})$ are obtained. In order to distinguish real video frames from synthesized ones, a discriminator $D(\cdot)$ extracts feature $y_i$ for the individual frame, and then a fusion function $\Theta$ would perform the temporal aggregation over all frames:
\begin{equation}
    y = \Theta(y_i) = \Theta(D(I_i)),~i=0,1,~\cdots,N_t-1.  \label{eq:2}
\end{equation}
The final discriminative logit $l$ is computed by a MLP $M$ conditioned on time steps:
\begin{equation}
    l = M(y, t_0, \cdots, t_{N_t-1}).
\end{equation}

In the following sections, we would make a comprehensive overhaul of the default config, leading to a new simple yet strong approach for video generation.

\begin{table}[t]
\footnotesize
\caption{
    \textbf{Evaluation of various configurations.}
    \textbf{Config-A} is the baseline model StyleGAN-V~\citep{skorokhodov2022stylegan}.
    \textbf{Config-B} and \textbf{Config-C} target fixing the texture sticking issue yet maintain the per-frame quality (\cref{subsec:alias_free_and_pretrain}), \rebuttal{which is primarily evaluated by FID.}
    \textbf{Config-D} aims to help the generator to produce more reasonable dynamics (\cref{subsec:tsm}), \rebuttal{which is primarily evaluated by FVD$_{16}$ and FVD$_{128}$.}
    \textbf{Config-E} and \textbf{Config-F} alleviate the discontinuity when interpolating motion embeddings (\cref{subsec:gobeyond_limit}).
    For all metrics, a lower number is better.
}
\label{tab:ablation}
\vspace{2pt}
\centering
\begin{tabular}{@{}llllllllll@{}}
\toprule
\multirow{2}{*}{\textbf{Configuration}}              & \multicolumn{3}{c}{SkyTimelapse} & \multicolumn{3}{c}{YouTube Driving} & \multicolumn{3}{c}{\rebuttal{Taichi-HD}} \\ \cmidrule(l){2-10} 
                                            & FID     & FVD$_{16}$     & FVD$_{128}$    & FID      & FVD$_{16}$      & FVD$_{128}$ & \rebuttal{FID}      & \rebuttal{FVD$_{16}$}      & \rebuttal{FVD$_{128}$}      \\ \midrule
\textbf{A}  StyleGAN-V                       & \textbf{40.8}  &   73.9   &    248.3        &   28.3 & 449.8        & 460.6      &    \rebuttal{33.8} & \rebuttal{152.0}  & \rebuttal{267.3}                  \\ \midrule
\textbf{B} + Alias free                            &  54.0    &  118.8    &   221.4    &   56.4       & 729.8           &       886.0   &  \rebuttal{31.4} &  \rebuttal{171.7} &  \rebuttal{522.9}  \\
\textbf{C} + Image pretrain                         &     52.2    & 73.5           &     230.3     &    \textbf{15.6}     &    272.8        &      447.5   &\rebuttal{\textbf{20.8}} & \rebuttal{104.3}  & \rebuttal{314.2}     \\ 
\textbf{D} + TSM      &  49.9    &  \textbf{49.0}       &     \textbf{135.9}       &    19.2      &      \textbf{207.2}      &         \textbf{221.5}  & \rebuttal{26.0} & \rebuttal{84.6} & \rebuttal{176.2}  \\
\textbf{E} + B-Spline    &    63.5   &    64.8    & 185.3   &     21.8     &    281.6        &     375.2  & \rebuttal{25.5}  & \rebuttal{\textbf{82.6}}  & \rebuttal{\textbf{169.7}}       \\
\textbf{F} + Low-rank  &     60.4    &    61.9       &     229.1       &    23.0      &     260.4       &   278.9   & \rebuttal{24.8} & \rebuttal{98.3}  & \rebuttal{186.8}      \\ \bottomrule
\end{tabular}
\vspace{-8pt}
\end{table}

\subsection{Alias-free operations and pre-learned knowledge}\label{subsec:alias_free_and_pretrain}

\begin{figure}[t]
\centering
\includegraphics[width=1.0\linewidth]{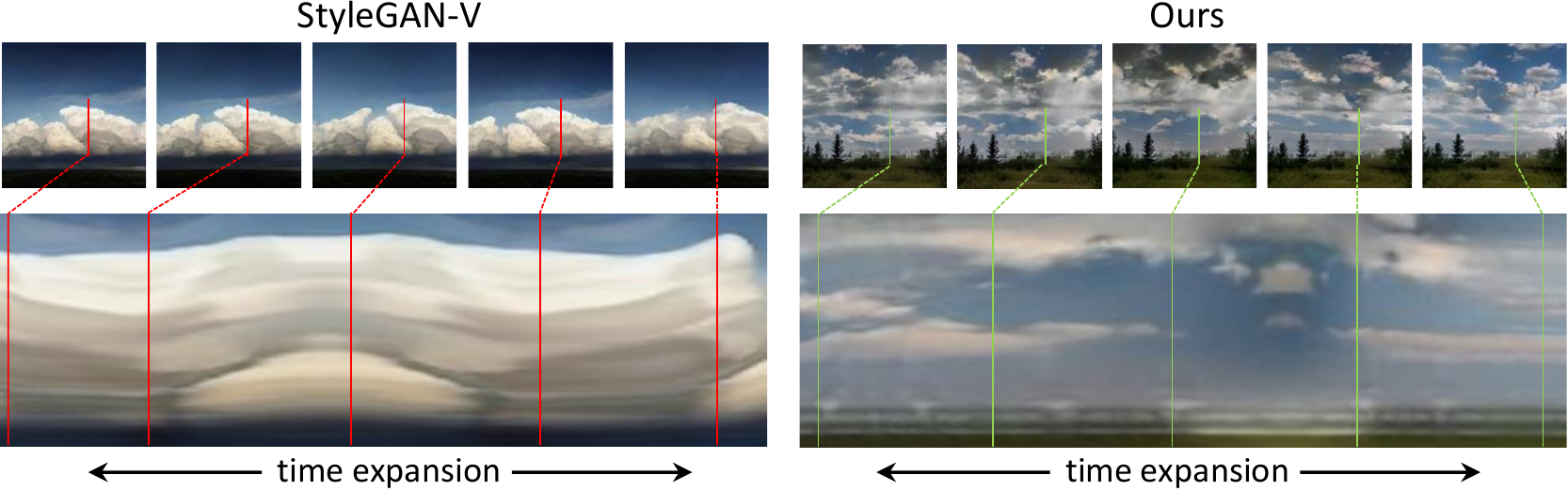}
\vspace{-18pt}
\caption{
    \textbf{Visualization of the texture sticking effect.}
    On the top are a few samples from a continuous sequence of frames, while the bottom visualizes the varying content at a fixed position.
    We can observe the brush effect from the baseline~\citep{skorokhodov2022stylegan}, where the pixel values are strongly anchored to the coordinates.
    Instead, our approach achieves smoother frame transition between neighbor frames.
}
\label{figure:sticking}
\vspace{0pt}
\end{figure}

%When composing a few frames, 
Generated frames within a short temporal window usually share similar visual concepts while the micro motion gradually occurs. Therefore, the smooth transition between frames tends to make the synthesis much more realistic. However, by examining the synthesized frames of StyleGAN-V, we find that texture sticking~\citep{karras2021alias} exists. 
We track the pixels at certain coordinates as the video continues and the \textit{brush effect} in \cref{figure:sticking}a indicates that these pixels actually move little. Thus, the texture sticks to fixed coordinates.

\textbf{Alias-free operations (Config-B).}
To mitigate the texture sticking issue, we follow the StyleGAN3~\citep{karras2021alias} to adopt the alias-free technique in the generator, which reduces the dependency on absolute pixel coordinates. 
As shown in \cref{figure:sticking}b, \textit{brush effect} disappears \emph{i.e.,} texture sticking is significantly alleviated by the alias-free technique. 

However, as reported in \cref{tab:ablation}, direct adoption of StyleGAN3 brings lower performance in terms of the FID and FVD scores. One possible reason is that the training receipt is directly inherited from the original StyleGAN-V without any modification. Despite its impairment on FID and FVD to some extent, we still use StyleGAN3 for the sake of its alias-free benefit.

\textbf{Pre-learned knowledge (Config-C).}
Simply replacing the generator backbone with StyleGAN3 deteriorates the synthesis. To fix it, we propose to make the best of pre-learned knowledge of image generation network.
As discussed before, the key challenge for video generation is to model temporal relations between frames. However, it is entangled with the target of modeling image distribution. We confirm that pre-training at the image level and then fine-tuning at the video level can decouple the two targets of modeling individual frames and cross-frame dynamics. As discovered in \citep{stylegan, yang2021semantic}, GAN's deep layers mainly control details, for example, coloring, detailed refinements, and etc. We hypothesize that these parts' knowledge can be pre-learned and re-used in the video generation learning process.

To be specific, we set $N_t$ as 1 in \cref{eq:2}, and train the whole network from scratch as the image pre-training stage. 
Then deep layers' weights are loaded in the video fine-tuning stage. With pre-trained deep layers' weights, now the optimization process can focus on early layers to learn to model sequential changes in structures that show realistic movements.
As shown in \cref{tab:ablation}, after adopting \textbf{Config-C}, the generated quality gets improved compared to \textbf{Config-B}. Now, with the alias-free technique, we achieve smooth frame transition \rebuttal{without compromising the per-frame quality.}

\subsection{Explicit Temporal Reasoning in Discriminator}\label{subsec:tsm}

In the adversarial learning of GAN, it is essential to have a strong discriminator, assuring the sufficient training of the generator. In the video generation task, the discriminator thus has to model the temporal relations of multiple frames to distinguish unnatural movements from real ones. However, in previous work only a simple concatenation operation $\oplus$ is used:
$y = \mathop{\oplus}\limits_{i}y_i$,
where $y_i$ denotes a single frame's feature extracted by the discriminator and $y$ denotes the feature after temporal fusion. 

\textbf{Temporal modeling (Config-D).} We incorporate an explicit temporal modeling approach \emph{i.e.,} temporal shift module (TSM) that shows exceptional performance for video classification.
Concretely, \rebuttal{at each convolution layer} before temporal fusion, we have features from multiple frames: \rebuttal{$y^n_i$} $\in \mathcal{R}^{H\times W\times C}, i=0,\cdots,N_{t}-1$, where \rebuttal{$n$ is the layer index}, $H$ and $W$ are the feature map resolution and $C$ is the feature channel number. Then, TSM performs channel-wise swapping between adjacent frames:
\rebuttal{
\begin{equation}
    y^n_i(x,y) \leftarrow \oplus_3 (y^n_{i-1}(x,y)[:\frac{C}{8}], y^n_i(x,y)[\frac{C}8:\frac{7C}8], y^n_{i+1}(x,y)[\frac{7C}8:]),
\end{equation}
}
where $\oplus_3(\cdot)$ is a concatenation operation, and $(x, y)$ are arbitrary coordinates.
In this way, one quarter of the channels of a single frame after temporal fusion contain information from adjacent frames. Follow-up convolution kernels in deeper layers can perform efficient temporal reasoning based on this mixed representation.
In our ablation study, while FID becomes slightly worse compared to \textbf{Config-C}, adopting explicit temporal modeling (\textbf{Config-D}) improves FVD$_{16}$ and FVD$_{128}$ substantially, yielding over 100\% improvement on YouTube Driving dataset. %This proves that adopting temporal clues in discriminator largely boost generator performance in mid-range temporal level.

\subsection{Towards Infinite Video Generation}\label{subsec:gobeyond_limit}
\begin{figure}[t]
\centering
\includegraphics[width=1.0\linewidth]{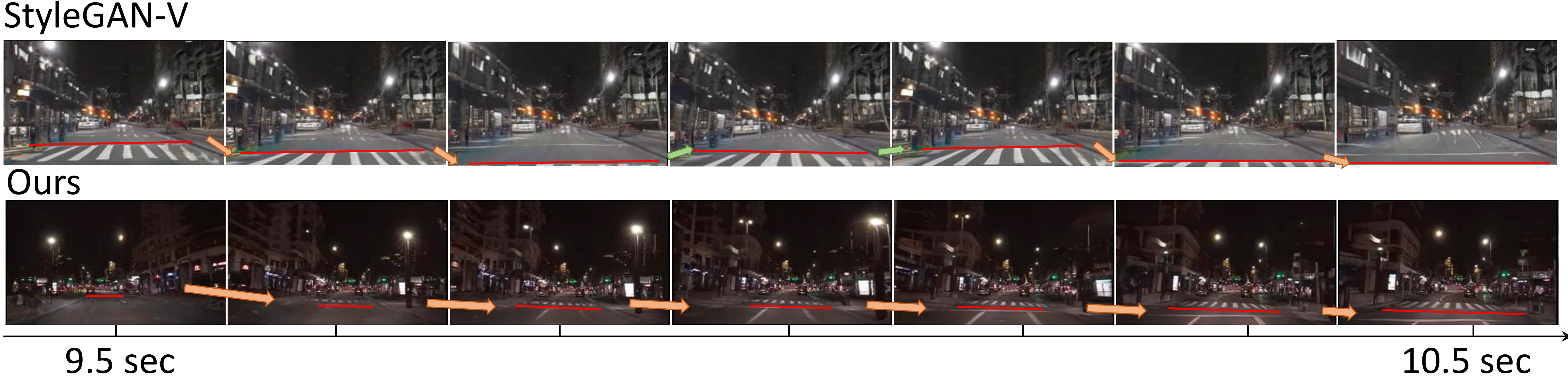}
\vspace{-18pt}
\caption{
    \textbf{Visualization of the content jittering effect.}
    As the video continues, we can observe the unstable crossroad in the baseline model~\citep{skorokhodov2022stylegan}, caused by discontinuous motion representation.
    Instead, our approach alleviates it through the proposed B-Spline motion representation.
}
\label{figure:jitter}
\vspace{-4pt}
\end{figure}

The metrics FID, FVD$_{16}$ and FVD$_{128}$ used before merely measure the synthesis quality of the relatively short videos (\emph{e.g.,} 128 frames usually cover around 6 seconds), we further investigate whether current configuration is able to produce arbitrarily long videos. 
\rebuttal{Note that, by saying arbitrarily long, we mean infinite length along the time axis, not infinite content.}
Ideally, we could easily generate an infinite number of frames by continuously sampling time $t$. The generator is supposed to produce the corresponding synthesis. However, as the video continues, a conspicuous jittering effect occurs periodically every 10 seconds.
%We continuously sample time $t$, rollouting the generated video, and turns out to find a conspicuous jittering phenomenon occurring periodically every 10 seconds.
For example, \cref{figure:jitter} shows the crossroad moves forward at the beginning, and then suddenly goes backward.

%motion. Before diving into this strange phenomenon, let us recap the design of motion embedding in StyleGAN-V.

\begin{figure}[t]
\centering
\includegraphics[width=0.8\linewidth]{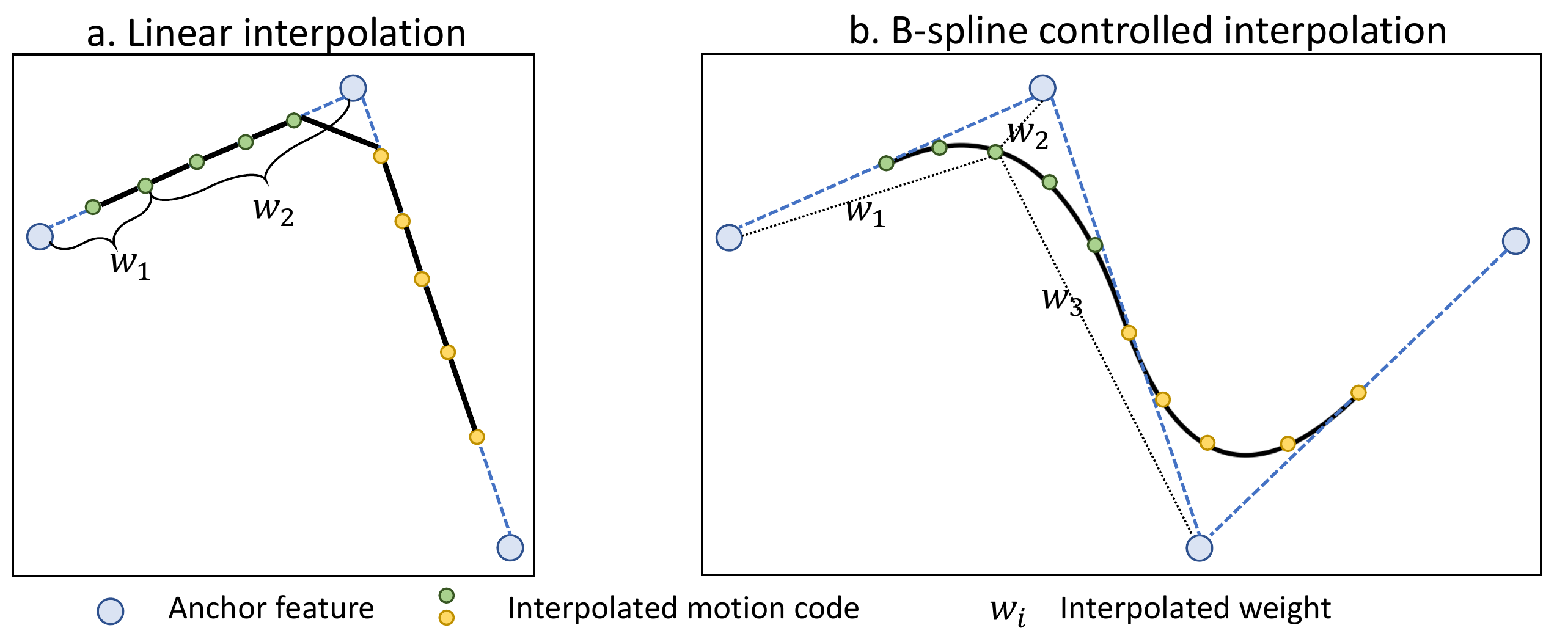}
\vspace{-5pt}
\caption{
    \textbf{Comparison between linear interpolation and our B-Spline controlled interpolation.}
    B-spline is able to smooth the interpolation between various anchor features, and hence improve the first-order numerical continuity.
}
\label{figure:bspline}
\vspace{0pt}
\end{figure}

\begin{figure}[t]
\centering
\includegraphics[width=0.8\linewidth]{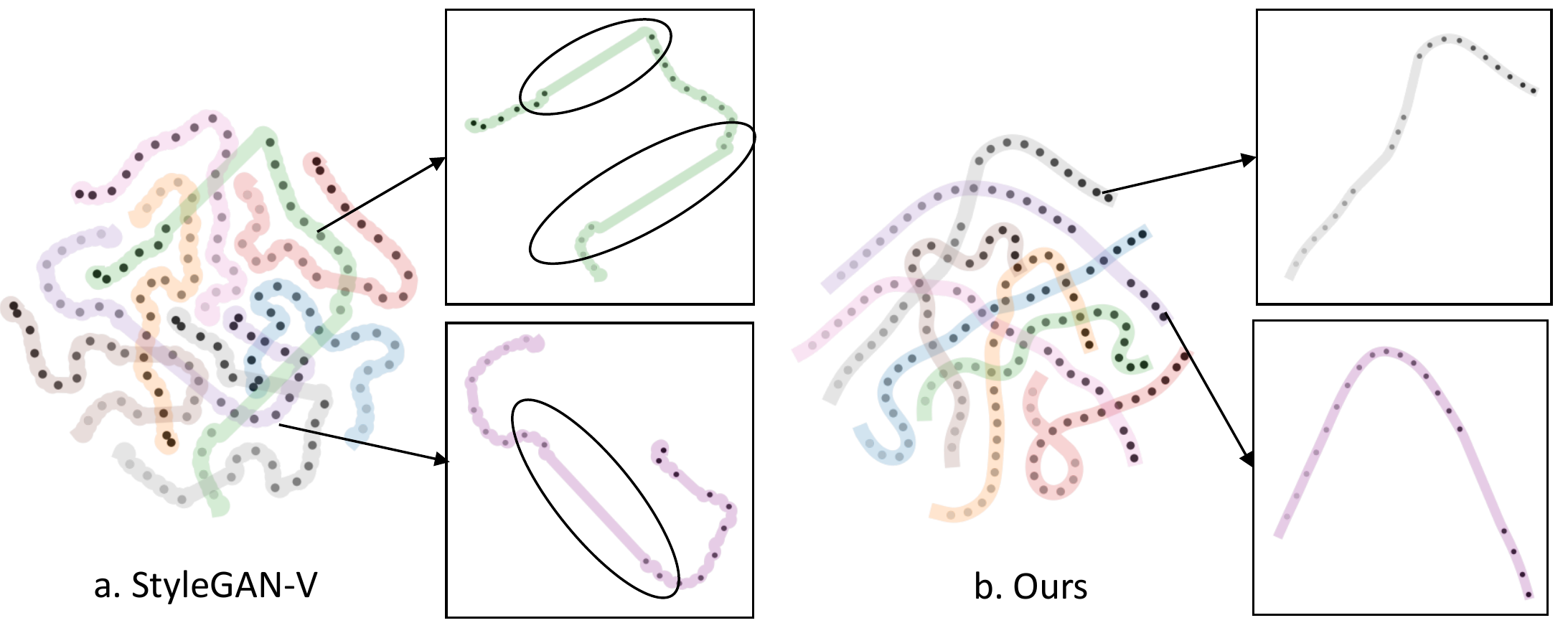}
\vspace{-5pt}
\caption{
    \textbf{Visualization of motion embeddings} using T-SNE~\citep{van2008visualizing}.
    Each dot refers to a motion code $v_t$, which represents a single frame's motion.
    A set of progressively darkening dots construct one trajectory that represents a synthesized frame sequence.
    We can tell that there usually exist undesired twists in the motion trajectory of StyleGAN-V~\citep{skorokhodov2022stylegan}, indicating the discontinuity of motion embedding.
    On the contrary, our approach helps alleviate this issue, resulting in smoother transition throughout the entire video.
}
\label{figure:motionemb}
\vspace{-15pt}
\end{figure}

\textbf{Discontinuity of motion embeddings.} 
As introduced in \cref{subsec:preliminary}, motion embedding $v_t$ contains time positional embedding $v_t^{pe}$ and interpolated motion embedding $v_t^{me}$. In particular, $v_t^{pe}$ could be obtained by a learned wave function on the left nearest anchor feature $a_l$ while the motion embedding $v_t^{me}$ derives from the linear interpolation between the left and right anchor feature $a_l$ and $a_{l+1}$. This interpolation would result in the first-order discontinuity for both learned wave function and linear interpolation when getting through multiple anchors, as shown in \cref{figure:bspline}a. 
Moreover, when the T-SNE~\citep{van2008visualizing} is applied to the motion embedding given a long period, we can see in \cref{figure:motionemb}a that such discontinuity causes the drastic change and jittering in the generated content.

\textbf{B-Spline based motion representations (Config-E).}
With an assumption of the discontinuity of motion embeddings, we design a basic spline (B-Spline) based motion representation that could guarantee first-order numerical smoothness. To be specific, B-Spline of order $n$ is a piece-wise polynomial function of degree $n-1$ in a given variable. It is widely used in computer-aided design and computer graphics for curve fitting, controlling a smooth curve with calculated weights $w_i$ on several control points. In particular, the number of control points is determined by the order of the B-Spline.
%We hypothesize that the jittering phenomenon is incurred by motion embedding's discontinuity.
%
%To alleviate it, we design a basic spline (B Spline) based motion embedding which guarantees first-order numerical smoothness.
%
%B Spline of order $n$ is a piece-wise polynomial function of degree $n-1$ in a given variable. It is widely used in computer-aided design and computer graphics for curve fitting. 
%
%It controls a smooth curve with calcuated weights $w_i$ on several control points.
%
%The number of control points is determined by the order of the B Spline.
%
With a pre-defined knot sequence $\{t_i\}$, a B-Spline with any order can be defined by means of the Cox–de Boor recursion formula~\citep{de1971subroutine}. The first-order B-Spline is then defined by:
\begin{equation}
    B_{i, 1}(t):= \begin{cases}1 & \text { if } t_i \leq t <t_{i+1}, \\ 0 & \text { otherwise. }\end{cases}
\end{equation}
Here $i$ is the interval index. B-spline of higher orders is further defined recursively:
\begin{equation}
    B_{i, k+1}(t):=\omega_{i, k}(t) B_{i, k}(t)+\left[1-\omega_{i+1, k}(t)\right] B_{i+1, k}(t),
\end{equation}
where $k$ is the order and
\begin{equation}
    \omega_{i, k}(t):= \begin{cases}\frac{t-t_i}{t_{i+k}-t_i} & t_{i+k} \neq t_i, \\ 0 & \text { otherwise. }\end{cases}
\end{equation}

We treat each anchor feature $a_i$ as the control point and use $B_{i,k}(t)$ as its corresponding weight given time $t$. Hence, B-Spline based anchor feature is defined as:
\begin{equation}
    \hat{a}(t) = \sum \limits_{i} B_{i,k}(t) a_i.
\end{equation}
We further use $\hat{a}(t)$ instead of discretized $a_i$ to calculate time positional embedding $v_t^{pe}$ and interpolated motion embedding $v_t^{me}$ as defined in \cref{eq:pe} and \cref{eq:me}.
% Practically we find that a B spline with an order of four effectively alleviates the discontinuity of motion embedding.
% %
% As visualized in \cref{figure:motionemb}b, the B Spline based motion embeddings traverse smoothly without obvious discontinuity compared to those by StyleGAN-V.
\cref{figure:bspline}b suggests the continuity property of B-Spline. Meanwhile, through T-SNE again, the B-Spline based motion representations become much smoother in \cref{figure:motionemb}b. As shown in \cref{figure:jitter}, the crossroad gradually approaches the observer without any sudden jittering. 

\textbf{Low-rank temporal modulation (Config-F).}
After adopting B-Spline based motion embedding, the jittering effect gets erased. However, we find that similar content periodically appears as video continues\rebuttal{(see \cref{figure:supp_repeat}a)}.
This result implies that motion embedding poorly represents visual concepts.
We thus hypothesize that the new motion embedding might be endowed with a stronger capacity to represent various visual concepts. 
Inspired by recent works~\citep{bau2020rewriting, wang2022rewriting}, we suppress the representation capacity through the low-rank strategy.
%Actually, with B Spline gauranteeing smoothness, motion embedding is endowed with stronger capacity, which should be suppressed properly.
%
%Inspired by recent works on fine-tuning generative models, We adopt a low-rank strategy~\tocite{zjy's siggraph} to suppress the over-strong representation capacity of the motion embedding.

Original StyleGAN-based architecture incorporates styles into convolutional kernels via modulation trick~\citep{karras2020analyzing}. When generating videos, a similar technique is also applied with small modifications. Content embedding $u$ together with the motion embedding $v_t$ would be fed into an affine layer $M$, generating the style code to modulate the original weight $W$:
\begin{equation}
   W^{\prime} = W\cdot M(u\oplus v_t), 
\end{equation}
where $\oplus$ stands for the concatenation operation. That is, motion and content embedding could equally contribute to the final style to change the visual concepts of frames. 
% Originally, to calculate the style which affects the distribution of each channel of the convolutional kernel $W$, content embedding $u$ and motion embedding $v_t$ are concatenated and mapped to the style embedding by an affine layer $M$: 
% \begin{equation}
%   f = W\cdot M(u\oplus v_t), 
% \end{equation}
% %
% , where $f$ stands for the feature after the modulated convolution. In this case, motion and content embedding serve symmetrically.
%
%
In order to suppress the representational capacity of motion embeddings, we first separate the motion and content codes and then replace the kernel regarding the motion code with a low-rank one:
\begin{equation}
\begin{aligned}
   W^{\prime} = W_{co}\cdot M_{co}(u) + W_{mo}\cdot M_{mo}(v_t), \text{ where }
    W_{mo}=UV,
     %&= W_{co}\cdot M_{co}(u) + UV\cdot M_{mo}(v_t),
\end{aligned}
\end{equation}
where $U, V$ is tall matrix that guarantees $W_{mo}$ is a low-rank matrix. 
With such a low-rank strategy, the capacity of motion embedding is suppressed. The repeating issue is alleviated in \cref{figure:supp_repeat}b.

% To suppress the motion embedding's representation capacity, 
% we first disengtangle motion and content embedding through separate kernels $W_{mo}$ and $W_{co}$:
% \begin{equation}
% \begin{aligned}
%   f &= W_{co}\cdot M_{co}(u) + W_{mo}\cdot M_{mo}(v_t) \\
%      &= W_{co}\cdot M_{co}(u) + UV\cdot M_{mo}(v_t),
% \end{aligned}
% \end{equation}
% , where $U, V$ is tall matrix which gaurantees that $W_{mo}=UV$ is a low rank kernel.
% %
% The low rank property of the kernel can suppress the capacity of motion embedding.
% %
% We find that with this low-rank time modulation, the cyclic repeatness over generated contents gets alleviated, as can be seen in our supplementary video.

\rebuttal{To summarize,  \textbf{Config-D} is the best choice for generating short video (less than 10 seconds), while \textbf{Config-F} is better for long video generation.}

%% file: sections/3.exp.tex
\section{Experiments}\label{sec:exp}

We evaluate our method on several datasets and compare it against prior arts.

\textbf{Evaluation Metrics.}
Akin to the literature, Frechet Inception Distance (FID)~\citep{heusel2017gans} and Frechet Video Distance (FVD)~\citep{unterthiner2018towards} serve as the quantitative metrics to evaluate the synthesis for image and video respectively. In terms of FVD, two temporal spans (consecutive 16 and 128 frames) are chosen to measure the synthesized videos.  

\textbf{Benchmarks.}
In Sec.~\ref{sec:method}, we have already conducted comprehensive ablation studies on SkyTimelapse~\citep{xiong2018learning}, \rebuttal{Taichi-HD~\citep{Siarohin_2019_NeurIPS}}, and YouTube Driving dataset~\citep{zhang2022learning}. To be specific, SkyTimelapse contains over two thousand videos which in average last twenty seconds. Various types of scenes (daytime, nightfall, and aurora) are included.  Similar to StyleGAN-V~\citep{skorokhodov2022stylegan}, we resize the videos of SkyTimelapse to $256 \times 256$ resolution. 
\rebuttal{Taichi-HD contains over three thousand videos recording a person performing Tai-chi. We also resize the videos to $256 \times 256$ resolution.}
YouTube Driving dataset~\citep{zhang2022learning} consists of 134 first-view driving videos with a total length of over 120 hours, covering up to 68 cities, showing various conditions, from different weather, different regions, to diverse scene types.
The driving scenes have tight geometrical constraints, with most of the objects (vehicles, buildings) following the rigid body constraint.
We resize the videos to $180\times 320$ resolution. 
A part of videos with countryside views in $320\times 640$ resolution is also chosen to benchmark high resolution video generation.
For training, we resize them to $256\times 256$ and $512\times 512$ resolution respectively.

Additionally, we compare our approach against previous methods including  MoCoGAN~\citep{tulyakov2018mocogan} and its StyleGAN2 based variant, MoCoGAN-HD~\citep{tian2021mocoganhd}, VideoGPT~\citep{yan2021videogpt}, DIGAN~\citep{yu2022dign}, TATS~\citep{ge2022long}, LongVideoGAN~\citep{brooks2022generating}, and StyleGAN-V~\citep{skorokhodov2022stylegan}.

\textbf{Training.}
We follow the training scheme of StyleGAN-V and train models on a server with 8 A100 GPUs.  
In terms of various methods and datasets, we grid search the $R_1$ regularization weight, whose details are available in Appendix. 
In particular, performances of MoCoGAN and its StyleGAN2 based variant, MoCoGAN-HD, VideoGPT, and DIGAN are directly borrowed from StyleGAN-V paper.

\begin{table}[t]
\centering\footnotesize
\setlength{\tabcolsep}{15pt}
\caption{
    \textbf{Quantitative comparison} between our approach and existing video generation methods on SkyTimelapse~\citep{xiong2018learning}.
    For all metrics, the smaller number the better.
}
\label{tab:sky}
\vspace{2pt}
\begin{tabular}{@{}llll@{}}
\toprule
       Method   & FID    & FVD$_{16}$    & FVD$_{128}$   \\ \midrule
MoCoGAN~\citep{tulyakov2018mocogan}              &  -        &    206.6       &  575.9        \\
~~+ StyleGAN2 backbone &  -        &    85.9       &    272.8      \\
MoCoGAN-HD~\citep{tian2021mocoganhd}           &  -         &   164.1        &  878.1        \\
VideoGPT~\citep{yan2021videogpt}             &  -         &   222.7        &  -        \\
DIGAN~\citep{yu2022dign}                &  -         &   83.1       &  196.7        \\
\rebuttal{LongVideoGAN}~\citep{brooks2022generating} & \rebuttal{-} & \rebuttal{116.5} & \rebuttal{152.7}                   \\
\rebuttal{TATS-base}~\citep{ge2022long}  & \rebuttal{-}  & \rebuttal{132.6}  & \rebuttal{-}  \\
StyleGAN-V~\citep{skorokhodov2022stylegan}           &   \textbf{40.8}     &     73.9      &  248.3        \\ \midrule
Ours                 &   49.9     & \textbf{49.0}    &      \textbf{135.9}    \\ \bottomrule
\end{tabular}
\vspace{-15pt}
\end{table}

\begin{table}[t]
\centering\footnotesize
\setlength{\tabcolsep}{15pt}
\caption{
    \rebuttal{
    \textbf{Quantitative comparison} between our approach and existing video generation methods on Taichi-HD~\citep{Siarohin_2019_NeurIPS}.
    For all metrics, a smaller number is better.
  } 
}
\label{tab:taichi}
\vspace{2pt}
\begin{tabular}{@{}llll@{}}
\toprule
Method   & FID    & FVD$_{16}$    & FVD$_{128}$   \\ \midrule
MoCoGAN-HD~\citep{tian2021mocoganhd}           &  -         &   144.7        &  -        \\
DIGAN~\citep{yu2022dign}                &  -         &   128.1       &  -      \\
TATS-base~\citep{ge2022long}  & -  & 94.6  & -  \\
StyleGAN-V~\citep{skorokhodov2022stylegan}           &   33.8     &   152.0      &  267.3       \\ \midrule
Ours                 &   \textbf{26.0}    & \textbf{84.6}    &      \textbf{176.2}    \\ \bottomrule
\end{tabular}
\vspace{-15pt}
\end{table}

\begin{table}[t]
\centering
\footnotesize
\setlength{\tabcolsep}{15pt}
\caption{
    \textbf{Quantitative comparison} between our approach and StyleGAN-V~\citep{skorokhodov2022stylegan} on the self-collected YouTube Driving dataset.
    For all metrics, a smaller number is better.
}
\label{tab:ytb}
\vspace{2pt}
\begin{tabular}{@{}cllll@{}}
\toprule
   Resolution        &  Method  & FID       & FVD$_{16}$      & FVD$_{128}$       \\ \midrule
\multirow{2}{*}{256} & StyleGAN-V &  28.3   &  449.8  & 460.6  \\
                     & Ours       &     \textbf{19.2}   &    \textbf{207.2}   &       \textbf{221.5}  \\ \midrule
\multirow{2}{*}{512} & StyleGAN-V &  14.6   &   262.4   &  285.4   \\
                     & Ours       &   \textbf{14.5}   &  \textbf{116.0}    &   \textbf{139.2}  \\ \bottomrule
\end{tabular}
\vspace{-15pt}
\end{table}

\subsection{Main Results}

As shown in \cref{tab:sky}, our method significantly outperforms existing baselines on SkyTimelapse in terms of FVD$_{16}$ and FVD$_{128}$.
\rebuttal{On Taichi-HD, our method achieves consistent improvement in terms of FID, FVD$_{16}$, and FVD$_{128}$ as reported in \cref{tab:taichi}.}
We also compare our method with StyleGAN-V in challenging YouTube Driving dataset.
at both $256\times 256$ and $512\times 512$ resolution.
For $256\times 256$ resolution in \cref{tab:ytb}, our method achieves better performance in both FID, FVD$_{16}$, and FVD$_{128}$.
For $512\times 512$ resolution, StyleGAN-V has a comparable FID score but much worse FVD score. For both two resolutions on YouTube driving, we yield a strong performance of less than 50\% FVD score than the prior art.
As shown in \cref{figure:gallery}, our method achieves an appealing visual effect, with smooth frame transition.

\begin{figure}
\centering
\vspace{-10pt}
\includegraphics[width=1.0\linewidth]{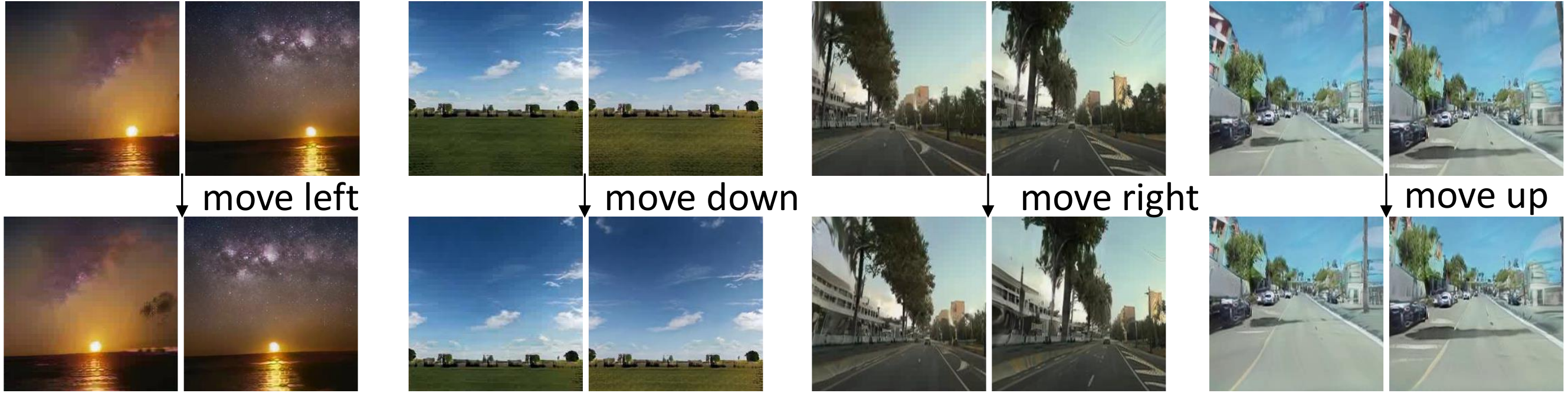}
\vspace{-18pt}
\caption{
    \textbf{Application of video 2D translation}, where we are able to control the shift of the frame content, such as moving the setting sun left and moving the grassland down.
}
\label{figure:eqt}
\vspace{0pt}
\end{figure}

\subsection{Strengths of the improved model}

\textbf{Good translation equivariance.}
We adopt alias-free technique from image generation for a smooth frame transition. 
This design also allows for geometric manipulation of the video using a human-specified translation matrix.
We showcase several examples in \cref{figure:eqt} by feeding different translation matrix to the video generator.
It is clear that videos can be successfully translated, and translation equivariance is well maintained over time.
This equivariance property improves human controllability over the video generation process.

\begin{wraptable}{r}{0.38\linewidth}
    \vspace{-20pt}
    \setlength{\tabcolsep}{3pt}
    % \begin{table}[h]
        \caption{FVD evaluated on the frames around anchors.}
        \label{tab:fidjit}
        \vspace{2pt}
        \centering
\begin{tabular}{cccc}
\toprule
& FVD$_{16}$                & FVD$_{16}$-anchor                \\ \midrule
\textbf{Config-D} &         207.2          &       269.1 (\textcolor{red}{+61.9})          \\
\textbf{Config-F} &        260.4           &       250.2 (\textcolor{green}{-10.2})          \\ \bottomrule
\end{tabular}
    \vspace{-8pt}
% \end{table}
\end{wraptable}
\noindent
\textbf{No jittering phenomenon.} 
As mentioned in \cref{subsec:gobeyond_limit}, discontinuity exists in the motion representation in the prior method, around which jittering phenomenon appears in generated videos.
It is caused by linear interpolation between discretized anchor features with fixed interval lengths (256 frames in practice).
Our method develops a novel B-Spline based motion representation to ensure the temporal smoothness over the anchor features. 
To quantify the smoothness of generated videos, we sample 16 frames across the anchor (eight frames before and behind the anchor) and calculate the FID$_{16}$ metric (which we name "FID$_{16}$-anchor") in YouTube Driving dataset.  
As reported in \cref{tab:fidjit}, \textbf{Config-D} suffers from severe performance drop when approaching anchors. While \textbf{Config-F} with B-Spline based motion representation maintains the performance, indicating that our method can generate long videos with smooth transitions without performance degrading.

%% file: sections/4.related_work.tex
\section{Related Work}\label{sec:relatedwork}

\noindent \textbf{Unconditional Video generation.}
% a. GAN
% b. diffusion
% c. autoregressive
% d. vae
% dvd gan, tats, yaohui wang, paper with code, 
Most of the video generation frameworks are built upon GANs~\citep{gan, biggan, stylegan, stylegan2, karras2021alias, pggan, dcgan}, owing to their flexibility as well as strong performance in image generation.
Prior arts \citep{tian2021mocoganhd, saito2017tgan, tulyakov2018mocogan, fox2021stylevideogan, skorokhodov2022stylegan, yu2022dign, vondrick2016generating, saito2020train, clark2019adversarial, wang2022latent} adopt image generators to synthesize frames from a time-coherent sequence of latent codes generated by recurrent networks. 
To explicitly disentangle the motion from the content, MoCoGAN~\citep{tulyakov2018mocogan} and TGAN~\citep{saito2017tgan} employ a motion code and a content code as input noise for the generator, serving as a common strategy for the later works.
StyleGAN-V~\citep{skorokhodov2022stylegan}, which we use as a strong baseline, and DIGAN~\citep{yu2022dign} both use implicit neural-based representations for continuous video synthesis.
\citet{brooks2022generating} leverages a multi-resolution training strategy to prioritize the time axis and ensure long-term consistency.
Another line of works ~\citep{yan2021videogpt, kalchbrenner2017videopixel, weissenborn2019scaling, rakhimov2020latent, walker2021prediction} use autoregressive models to achieve video generation. 
~\citet{kalchbrenner2017videopixel} use a PixelCNN~\citep{van2016pixelcnn} decoder to synthesize the next frame pixel by pixel in an autoregressive manner. 
VideoGPT~\citep{yan2021videogpt} built upon VQ-VAE adopts an autoregressive model to decompose videos into a sequence of tokens. 
TATS-base~\citep{ge2022long} similarly leverages 3D-VQGAN to decompose videos into tokens and further uses transformers to model the relationship between frame tokens. 
Additionally, some works using diffusion-based models also present promising results on video generation. 
Video diffusion~\citep{ho2022videodiffusion} models entire videos using a 3D video architecture while ~\citet{yang2022diffusion} uses an image diffusion model to synthesize frames with a recurrent network.
%
% However, they may suffer from synthesizing long videos with realistic and smooth motions. 
% %
% To alleviate these, we introduce several new components like a time-aware discriminator and better motion representations, which help us to achieve substantial improvements over baselines quantitatively and qualitatively.

\noindent \textbf{Video generation with guided conditions.}
A close line of research is video prediction, aiming to generate full videos from the observation of previous frames~\citep{babaeizadeh2017stochastic, kumar2019videoflow, lee2018stochastic, luc2020transformation, nash2022transframer, finn2016unsupervised} or a given conditions~\citep{Kim2020_GameGan, Kim2021_DriveGAN, ha2018worldmodels}. 
These works typically employ the reconstruction losses to make the future frames predictable for the model.
Video translation \citep{pan2019video, Yang_2018_ECCV, wang2018video, chan2019everybody, Siarohin_2019_CVPR, ren2020human, Siarohin_2019_NeurIPS} is also a paradigm for video synthesis, which translates the given segmentation masks and keypoints into videos.
Some~\citep{li2022infinitenature, liu2021infinite, ren2022look} focus on synthesizing videos of indoor and outdoor scenes for given camera trajectories, which need to model the 3D scene explicitly.
Besides, ~\citet{ho2022videodiffusion, hong2022cogvideo, wu2021godiva, wu2021n} explore text-to-video generation with diffusion-based models and transformers. 

%% file: sections/5.conclusion.tex
\section{Conclusion and Discussion}\label{sec:conclusion}
In this work, we set up a simple yet effective video generation baseline. 
By introducing the alias-free operations together with pre-learned knowledge, per-frame quality is improved. Explicit temporal modeling in discriminator enhances the dynamic synthesis. In addition, a new B-Spline based motion representation is proposed to enable smooth infinite frame generation. Experimental results show substantial improvements over the previous methods on different datasets. 

\textbf{Discussion.} Our approach has several limitations. The current framework follows prior arts using two latent codes to represent content and motion respectively. However, in our experiments, it is suggested that they usually entangle with each other, especially for the driving scenes where it is difficult to separate the content and motion.
% For instance, a moving car may occur as a part of dynamics while a static one belongs to the whole background, confusing the learning to some extent. That is, a clear learning objective might be  potential in helping the disentanglement in future. 
\rebuttal{Moreover, the frame quality is still far from ideal. The object shape is not consistent across consecutive frames, producing apparent artifacts. This could be further improved by introducing structural priors.
Last, although we could generate infinite videos by gradually changing content code, direct control of novel content over time is lacking. 
\citet{ge2022long} uses explicit planning in the latent space to achieve the controllability of the synthesis, especially for infinite video generation. Similar ideas could be adopted in our framework to improve the quality of long videos.} 

%% file: sections/6.ref.tex
\bibliography{ref}
\bibliographystyle{iclr2023_conference}

%% file: sections/7.appendix.tex
\newpage
\appendix
\newcommand{\AppendixPrefix}{A}
\renewcommand{\thefigure}{\AppendixPrefix\arabic{figure}}
\setcounter{figure}{0}
\renewcommand{\thetable}{\AppendixPrefix\arabic{table}} 
\setcounter{table}{0}
\renewcommand{\theequation}{\AppendixPrefix\arabic{equation}} 
\setcounter{equation}{0}

\section*{Appendix}

This appendix is organized as follows. \cref{sec:implementations} and \cref{sec:datasets} present the implementation and dataset details respectively, followed by \cref{sec:userstudy} and \cref{sec:more_results} showing user study result and more visual results.

\section{Implementation Details.}\label{sec:implementations}
Our method is developed based on the official implementation of StyleGAN-V~\citep{skorokhodov2022stylegan}.
We adopt hyper-parameters, the optimizer, the loss function, and the training script to ensure a fair comparison.
Notably, the total number of seen images for sufficient data training is 25 million regardless of datasets.
We evaluate once after training on every 2.5 million images, and report the result with highest FVD$_{16}$ score.

Due to statistics variance of different datasets, we search training regularization term, \emph{i.e.} $R_1$ value, for each method and dataset.
Empirically, we find that a smaller $R_1$ value (\emph{e.g.,} 0.25) works well for pretraining stage (\textbf{Config-C}). While a larger $R_1$ value (\emph{e.g.,} 4) better suits to video generation learning.

\section{Dataset Details.}\label{sec:datasets}

In this section, we introduce the datasets we use, briefly analyze their characteristics and challenges brought for video generation, and show sampled real video frames.

\textbf{SkyTimelapse.}
SkyTimelapse~\citep{xiong2018learning} is a time-lapse dataset collected from the Internet showing dynamic sky scenes,
such as the cloudy sky with moving clouds, and the starry sky with moving stars.
It contains various conditions, for instance, different weather conditions (daytime, nightfall, dawn,
starry night and aurora), different kinds of foreground object (cloudy sky, starry sky, and sunny sky), and different motion patterns of the sky. 
It also contains some other objects like trees, mountains, and buildings which further improve the visual diversity.
However, since clouds are fluids, there are very few constraints on movement. Minor disturbances and deformations can make a generated video look realistic.

\begin{figure}
\centering
\includegraphics[width=\linewidth]{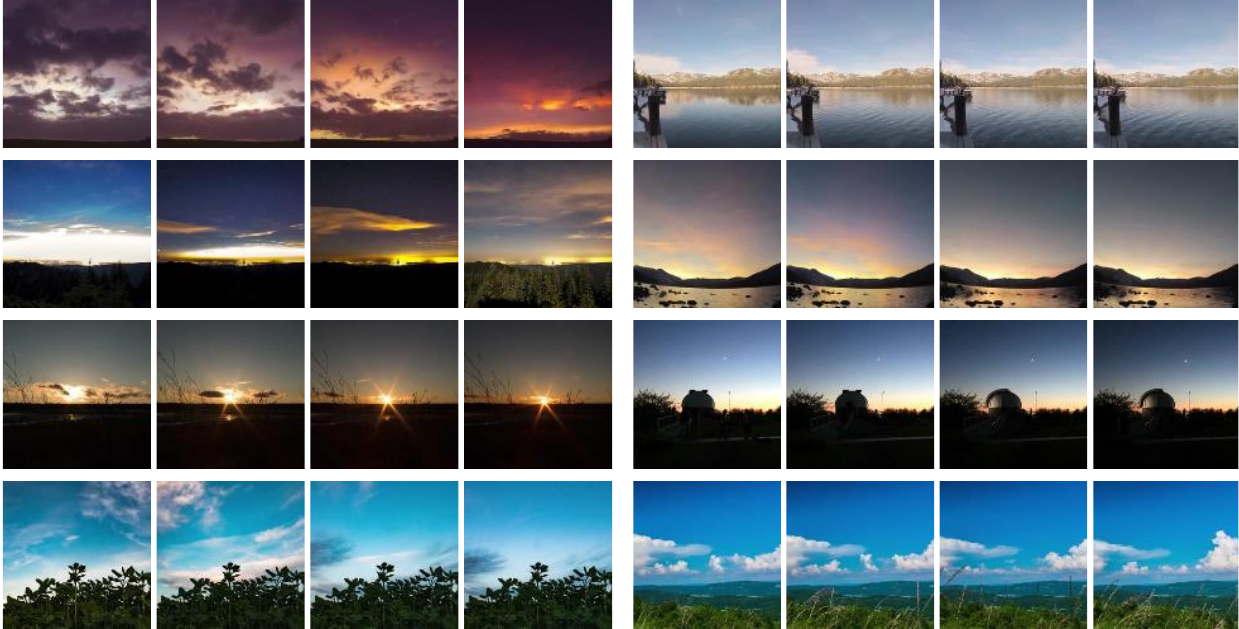}
\caption{
Sampled video frames from SkyTimelapse~\citep{xiong2018learning}.
}
\label{figure:supp_sky_gt}
\end{figure}

The whole dataset contains over two thousand videos which in average last twenty seconds. We resize the videos to $256\times 256$ following prior works. We sample several video clips in \cref{figure:supp_sky_gt}.

\begin{figure}
\centering
\includegraphics[width=\linewidth]{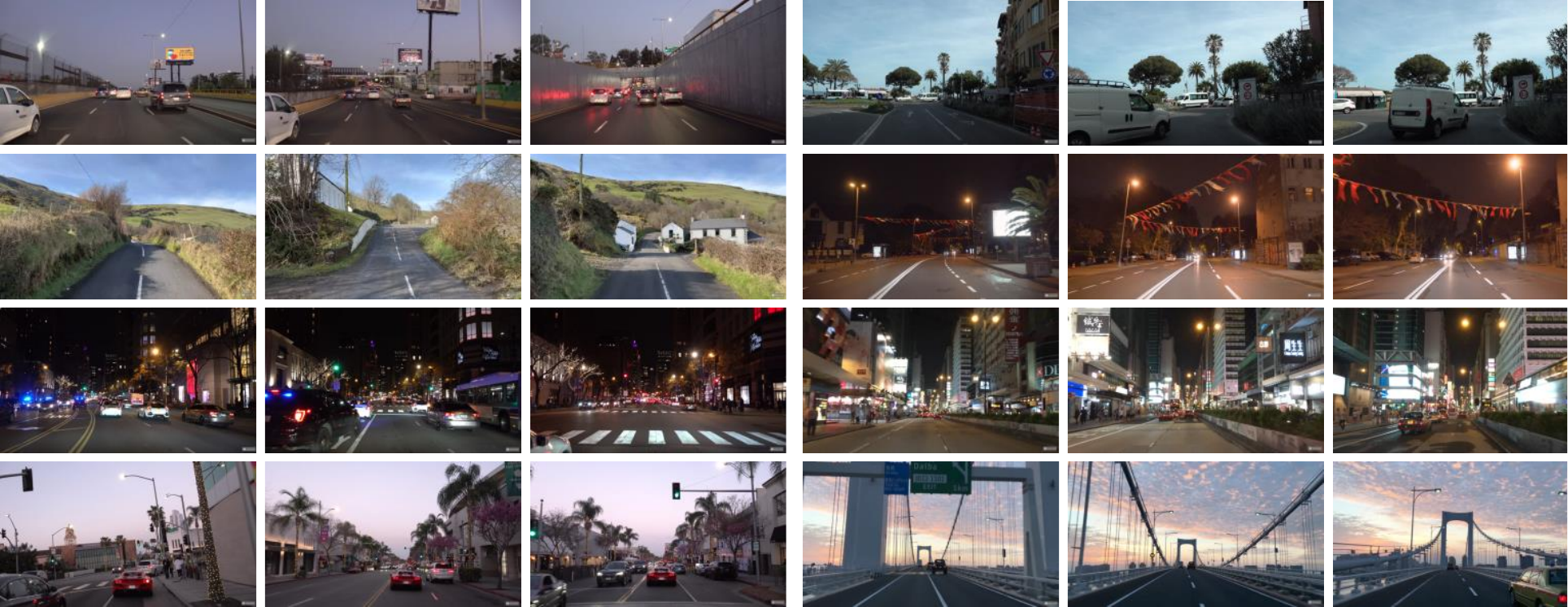}
\caption{
Sampled video frames from YouTube Driving dataset~\citep{zhang2022learning}.
}
\label{figure:supp_ytb_gt}
\end{figure}

\textbf{YouTube Driving.}
YouTube Driving Dataset~\citep{zhang2022learning} is crawled from YouTube which contains a massive amount of real-world driving frames with various conditions, from different weather, different regions, to diverse scene types.
To be specific, 134 videos with a total length of over 120 hours are collected, covering up to 68 cities. \cref{figure:supp_ytb_gt} shows sampled frames. The videos are resized to $256\times 256$ resolution for training.

The driving scenes have tighter geometrical constraints, with most of the objects (vehicles, buildings) following the rigid body constraint. The 3D structure of the whole scene should be maintained without any deformation.
Besides, different types of objects show different motion patterns. Roads and buildings recede at a reasonable speed opposite to that of ego car, while other cars show lane changes, stops, accelerations.

\begin{figure}
\centering
\includegraphics[width=\linewidth]{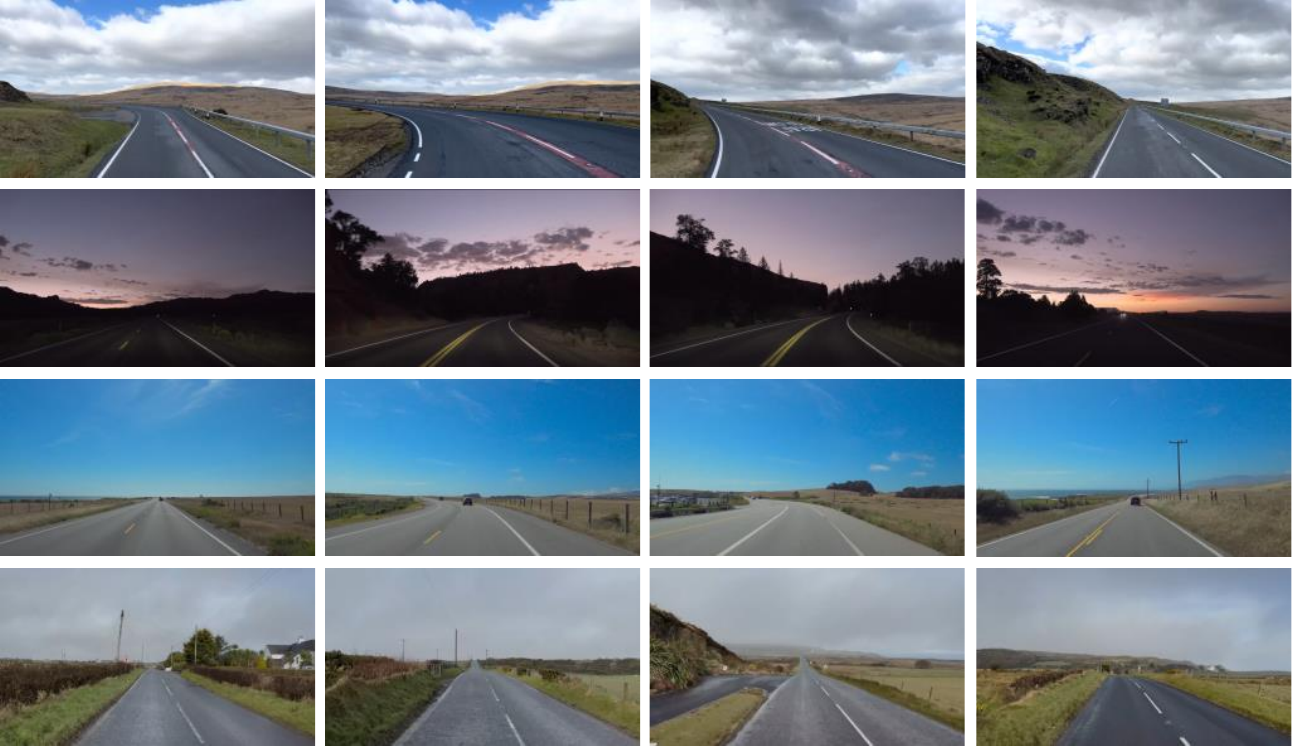}
\caption{
Sampled video frames from Countryside dataset~\citep{zhang2022learning}.
}
\label{figure:supp_cts_gt}
\end{figure}

\textbf{Countryside.}
We select a part of videos with countryside scenes from YouTube Driving dataset and resize them to $512\times 512$ resolution for training. We use this dataset to benchmark the performance of generating high-resolution videos. Real samples are presented in \cref{figure:supp_cts_gt}.

\color{revision_color}
\textbf{Taichi-HD.} Taichi-HD~\citep{Siarohin_2019_NeurIPS} is a collection of YouTube videos recording a single person performing Tai-chi. 
It contains 3049 training videos in total, with various foreground objects (people with different styles of clothing) and diverse background.
It requires realistic human motion generation, which contains non-rigid motion modeling, which poses a significant challenge for video generation.

\section{User Study}\label{sec:userstudy}
\begin{table}[t]
\centering
\footnotesize
\caption{\rebuttal{\textbf{User study result.} We report the percentage of favorite users for the quality of a set of single images, a set of short videos, and a set of long videos under models from three different settings.}}
\label{supp:userstudy}
\begin{tabular}{@{}ccccccc@{}}
\toprule
         & \multicolumn{3}{c}{YouTube Driving} & \multicolumn{3}{c}{Taichi-HD}       \\ \midrule
         & image(\%)  & short video(\%)  & long video(\%)  & image(\%) & short video(\%) & long video(\%) \\ \midrule
Config-A &  4    &      6        &        8     &     0  &     2        &    4        \\
Config-D &   \textbf{62}     &    \textbf{60}          &     12        &    \textbf{52}   &       \textbf{66}      &   36         \\
Config-F &   34     &      34        &       \textbf{80}      &     48  &       32      &      \textbf{60}      \\ \bottomrule
\end{tabular}
\end{table}

To further prove our method's efficacy, we conduct user study over baseline (\textbf{Config-A}), \textbf{Config-D}, and \textbf{Config-F} on YouTube Driving dataset and Taichi-HD dataset.
We evaluate from three perspectives:

\noindent
\textbf{Single Image Quality.} We provide user with a collection of images randomly sampled from generated videos.

\noindent
\textbf{Short Video Quality.} We provide user with a collection of generated videos with consecutive 128 frames (up to about 5 seconds). 

\noindent
\textbf{Long Video Quality.} We provide user with a collection of generated videos with consecutive 500 frames (up to 20 seconds).

As shown in \cref{supp:userstudy}, in terms of single image quality, users prefer \textbf{Config-D} over other settings in both YouTube Driving and Taichi-HD dataset.
The same trend occurs in short video quality assessment. With pre-learned knowledge and a stronger discriminator with temporal shift module (TSM), \textbf{Config-D} reaps a significantly higher number of votes.
However, for long video generation, \textbf{Config-F} becomes the best choice for both YouTube Driving and Taichi-HD dataset. Unnatural jittering and discontinuous motion can drastically affect the users' video viewing experience. With B-Spline based motion embedding and low rank time modulation, \textbf{Config-F} is undoubtedly the best option for long video generation.

\color{black}

\section{More Visual Results.}\label{sec:more_results}

\begin{figure}
\centering
\includegraphics[width=\linewidth]{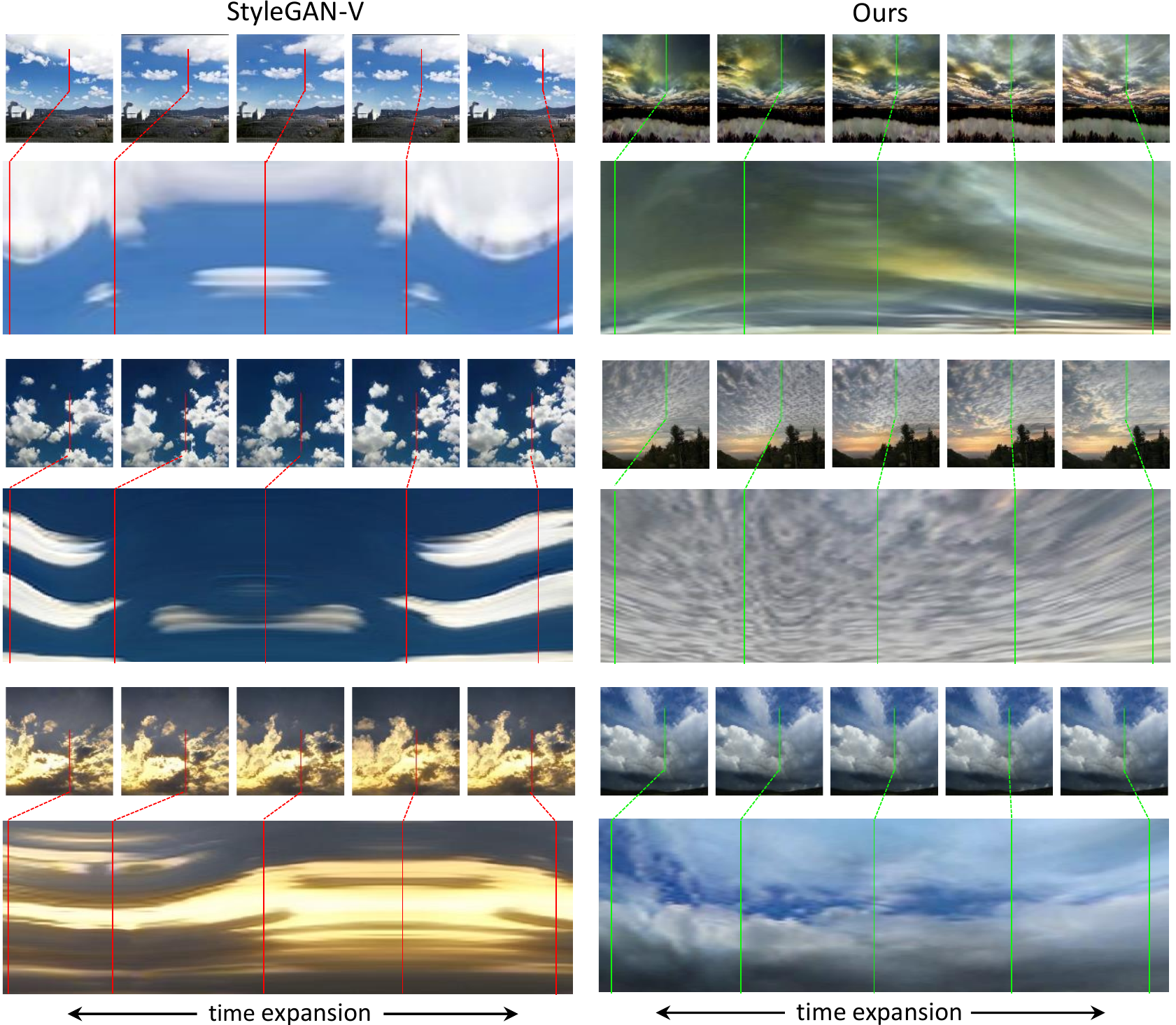}
\caption{
\textbf{More examples on texture sticking phenomenon.}  We concatenate slices in each frame of the generated video (marked by red and green). We observe \textit{brush effect} when texture sticks at specific location. Instead, our approach achieves more smooth frame transition between neighbor frames.
}
\label{figure:supp_texture_sticking}
\end{figure}

\textbf{Texture Sticking phenomenon.}
Texture sticking exists in videos generated by prior methods.
Our method leverages the alias-free technique in the generator to overcome this issue.
\cref{figure:supp_texture_sticking} shows more comparison between our method and StyleGAN-V~\citep{skorokhodov2022stylegan}.

\begin{figure}
\centering
\includegraphics[width=\linewidth]{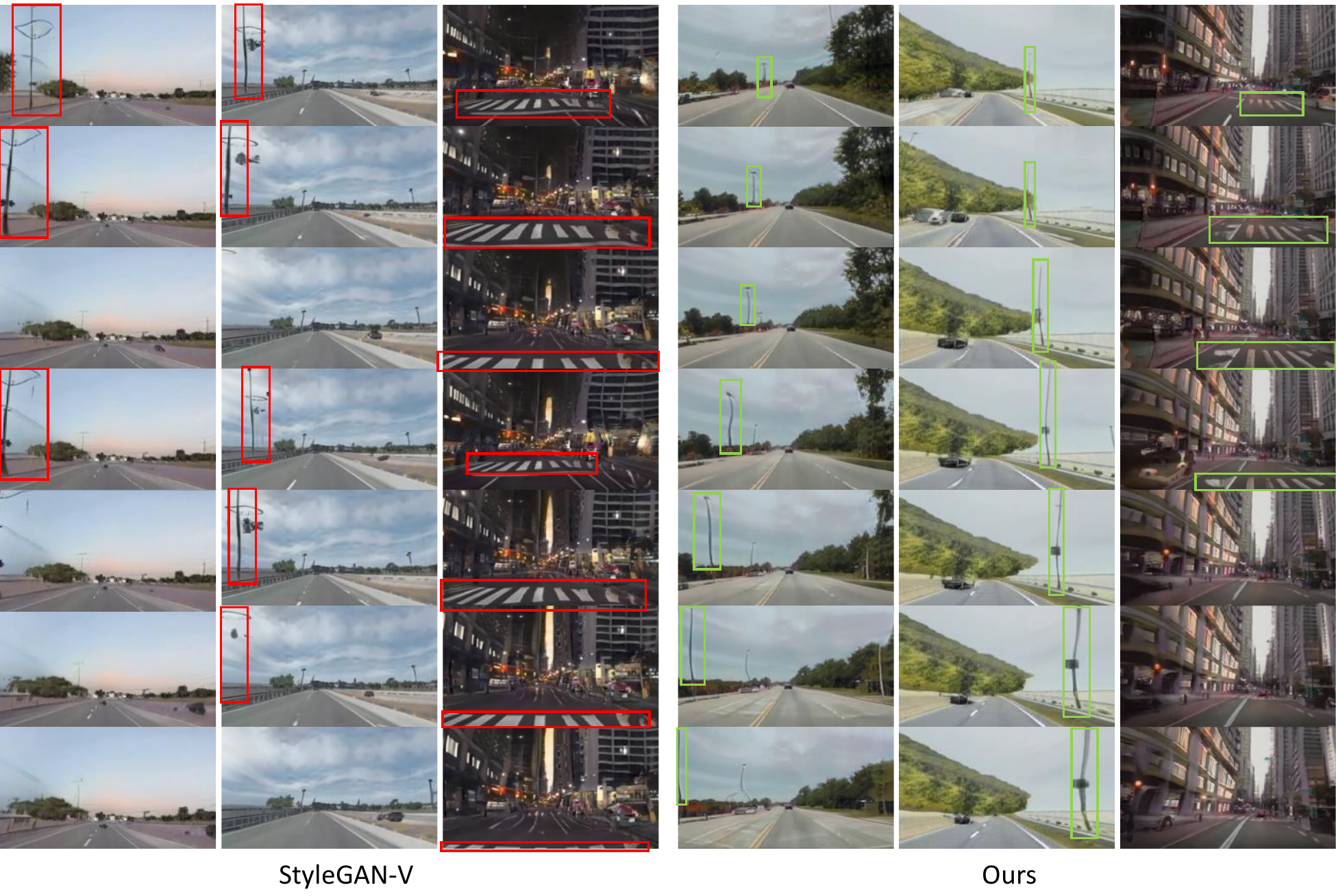}
\caption{
\textbf{More examples on jittering phenomenon.} Left three columns denote video frames produced by StyleGAN-V~\citep{skorokhodov2022stylegan}. Obviously, as highlighted in red bounding boxes, the street lights and crossroad move unsteadily (\emph{e.g.,} gradually fading out first and suddenly fading in). On the contrary, our approach enables much more reasonable and continuous dynamics of certain objects (see green bounding boxes). 
}
\label{figure:supp_jit}
\end{figure}

\textbf{Jittering phenomenon.}
Motion embedding designed in previous method is not first-order continuous, leading to jittering phenomenon when generating long videos.
We develop B-Spline based motion embedding which guarantees smoothness and alleviates the problem. 
We show examples comparing video sequences generated by our method with StyleGAN-V~\citep{skorokhodov2022stylegan} in \cref{figure:supp_jit}.
Our videos are smooth, without abnormal movements. In videos generated by StyleGAN-V~\citep{skorokhodov2022stylegan}, street lights and zebra crossings may suddenly change their moving directions.

\begin{figure}
\centering
\includegraphics[width=\linewidth]{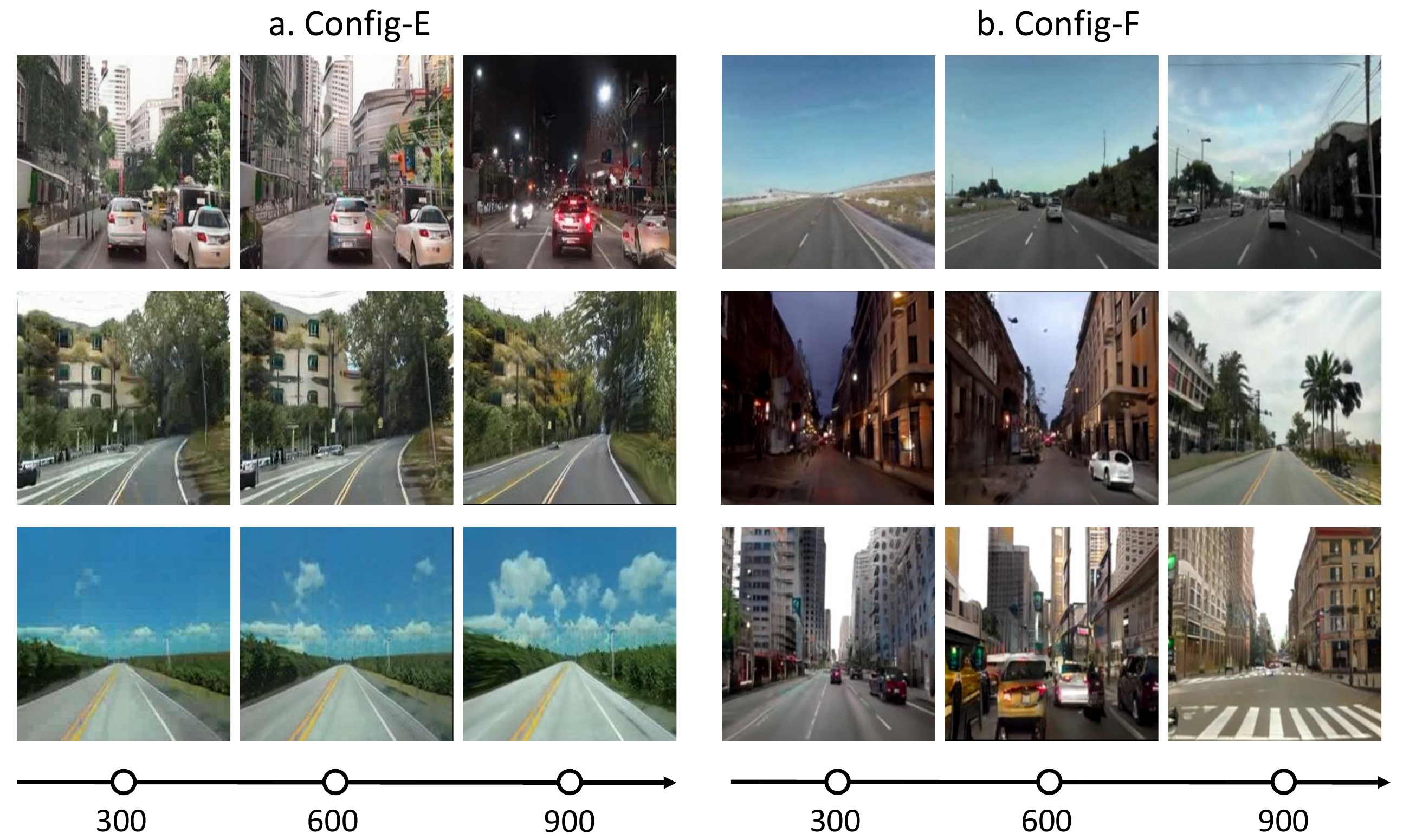}
\caption{
\rebuttal{\textbf{Examples on repeating contents.} Left three columns denote video frames produced by \textbf{Config-E} at the 300th, 600th, and 900th frame respectively. Similar contents (such as foreground objects, and scene layout) periodically appear. In contrast, after adopting low rank time modulation (\textbf{Config-F}), contents change gradually without repeating.} 
}
\label{figure:supp_repeat}
\end{figure}

\color{revision_color}
\textbf{Repeating contents.} With B-Spline based motion embedding, \textbf{Config-E} can compose smooth long videos. However, similar contents appear periodically as shown in \cref{figure:supp_repeat}a.
B-Spline based motion embedding is endowed with so strong capacity that it represents content and motion concepts simultaneously, rather than motion concepts only. We suppress the representation capacity through the low-rank strategy.
As shown in \cref{figure:supp_repeat}b, \textbf{Config-F} with low rank time modulation can generate progressively forward contents without repeating contents.
\color{black}

%% file: main.bbl
\begin{thebibliography}{65}
\providecommand{\natexlab}[1]{#1}
\providecommand{\url}[1]{\texttt{#1}}
\expandafter\ifx\csname urlstyle\endcsname\relax
  \providecommand{\doi}[1]{doi: #1}\else
  \providecommand{\doi}{doi: \begingroup \urlstyle{rm}\Url}\fi

\bibitem[Alaluf et~al.(2022)Alaluf, Patashnik, Wu, Zamir, Shechtman,
  Lischinski, and Cohen-Or]{alaluf2022third}
Yuval Alaluf, Or~Patashnik, Zongze Wu, Asif Zamir, Eli Shechtman, Dani
  Lischinski, and Daniel Cohen-Or.
\newblock Third time's the charm? image and video editing with stylegan3.
\newblock \emph{arXiv preprint arXiv:2201.13433}, 2022.

\bibitem[Babaeizadeh et~al.(2017)Babaeizadeh, Finn, Erhan, Campbell, and
  Levine]{babaeizadeh2017stochastic}
Mohammad Babaeizadeh, Chelsea Finn, Dumitru Erhan, Roy~H Campbell, and Sergey
  Levine.
\newblock Stochastic variational video prediction.
\newblock \emph{arXiv preprint arXiv:1710.11252}, 2017.

\bibitem[Bau et~al.(2020)Bau, Liu, Wang, Zhu, and Torralba]{bau2020rewriting}
David Bau, Steven Liu, Tongzhou Wang, Jun-Yan Zhu, and Antonio Torralba.
\newblock Rewriting a deep generative model.
\newblock In \emph{European conference on computer vision}, pp.\  351--369.
  Springer, 2020.

\bibitem[Brock et~al.(2019)Brock, Donahue, and Simonyan]{biggan}
Andrew Brock, Jeff Donahue, and Karen Simonyan.
\newblock Large scale {GAN} training for high fidelity natural image synthesis.
\newblock In \emph{Int. Conf. Learn. Represent.}, 2019.

\bibitem[Brooks et~al.(2022)Brooks, Hellsten, Aittala, Wang, Aila, Lehtinen,
  Liu, Efros, and Karras]{brooks2022generating}
Tim Brooks, Janne Hellsten, Miika Aittala, Ting-Chun Wang, Timo Aila, Jaakko
  Lehtinen, Ming-Yu Liu, Alexei~A Efros, and Tero Karras.
\newblock Generating long videos of dynamic scenes.
\newblock \emph{arXiv preprint arXiv:2206.03429}, 2022.

\bibitem[Chan et~al.(2019)Chan, Ginosar, Zhou, and Efros]{chan2019everybody}
Caroline Chan, Shiry Ginosar, Tinghui Zhou, and Alexei~A Efros.
\newblock Everybody dance now.
\newblock In \emph{Int. Conf. Comput. Vis.}, 2019.

\bibitem[Clark et~al.(2019)Clark, Donahue, and Simonyan]{clark2019adversarial}
Aidan Clark, Jeff Donahue, and Karen Simonyan.
\newblock Adversarial video generation on complex datasets.
\newblock \emph{arXiv preprint arXiv:1907.06571}, 2019.

\bibitem[de~Boor(1971)]{de1971subroutine}
Carl de~Boor.
\newblock Subroutine package for calculating with b-splines.
\newblock Technical report, Los Alamos National Lab.(LANL), Los Alamos, NM
  (United States), 1971.

\bibitem[Finn et~al.(2016)Finn, Goodfellow, and Levine]{finn2016unsupervised}
Chelsea Finn, Ian Goodfellow, and Sergey Levine.
\newblock Unsupervised learning for physical interaction through video
  prediction.
\newblock \emph{Adv. Neural Inform. Process. Syst.}, 2016.

\bibitem[Fox et~al.(2021)Fox, Tewari, Elgharib, and
  Theobalt]{fox2021stylevideogan}
Gereon Fox, Ayush Tewari, Mohamed Elgharib, and Christian Theobalt.
\newblock Stylevideogan: A temporal generative model using a pretrained
  stylegan.
\newblock \emph{arXiv preprint arXiv:2107.07224}, 2021.

\bibitem[Ge et~al.(2022)Ge, Hayes, Yang, Yin, Pang, Jacobs, Huang, and
  Parikh]{ge2022long}
Songwei Ge, Thomas Hayes, Harry Yang, Xi~Yin, Guan Pang, David Jacobs, Jia-Bin
  Huang, and Devi Parikh.
\newblock Long video generation with time-agnostic vqgan and time-sensitive
  transformer.
\newblock \emph{arXiv preprint arXiv:2204.03638}, 2022.

\bibitem[Goetschalckx et~al.(2019)Goetschalckx, Andonian, Oliva, and
  Isola]{goetschalckx2019ganalyze}
Lore Goetschalckx, Alex Andonian, Aude Oliva, and Phillip Isola.
\newblock Ganalyze: Toward visual definitions of cognitive image properties.
\newblock In \emph{Int. Conf. Comput. Vis.}, pp.\  5744--5753, 2019.

\bibitem[Goodfellow et~al.(2014)Goodfellow, Pouget-Abadie, Mirza, Xu,
  Warde-Farley, Ozair, Courville, and Bengio]{gan}
Ian Goodfellow, Jean Pouget-Abadie, Mehdi Mirza, Bing Xu, David Warde-Farley,
  Sherjil Ozair, Aaron Courville, and Yoshua Bengio.
\newblock Generative adversarial nets.
\newblock In \emph{Adv. Neural Inform. Process. Syst.}, 2014.

\bibitem[Ha \& Schmidhuber(2018)Ha and Schmidhuber]{ha2018worldmodels}
David Ha and J{\"u}rgen Schmidhuber.
\newblock Recurrent world models facilitate policy evolution.
\newblock In \emph{Adv. Neural Inform. Process. Syst.}, 2018.

\bibitem[Heusel et~al.(2017)Heusel, Ramsauer, Unterthiner, Nessler, and
  Hochreiter]{heusel2017gans}
Martin Heusel, Hubert Ramsauer, Thomas Unterthiner, Bernhard Nessler, and Sepp
  Hochreiter.
\newblock Gans trained by a two time-scale update rule converge to a local nash
  equilibrium.
\newblock \emph{Advances in neural information processing systems}, 30, 2017.

\bibitem[Ho et~al.(2022)Ho, Salimans, Gritsenko, Chan, Norouzi, and
  Fleet]{ho2022videodiffusion}
Jonathan Ho, Tim Salimans, Alexey Gritsenko, William Chan, Mohammad Norouzi,
  and David~J Fleet.
\newblock Video diffusion models.
\newblock \emph{arXiv preprint arXiv:2204.03458}, 2022.

\bibitem[Hong et~al.(2022)Hong, Ding, Zheng, Liu, and Tang]{hong2022cogvideo}
Wenyi Hong, Ming Ding, Wendi Zheng, Xinghan Liu, and Jie Tang.
\newblock Cogvideo: Large-scale pretraining for text-to-video generation via
  transformers.
\newblock \emph{arXiv preprint arXiv:2205.15868}, 2022.

\bibitem[Jahanian et~al.(2020)Jahanian, Chai, and
  Isola]{jahanian2019steerability}
Ali Jahanian, Lucy Chai, and Phillip Isola.
\newblock On the" steerability" of generative adversarial networks.
\newblock \emph{Int. Conf. Learn. Represent.}, 2020.

\bibitem[Kalchbrenner et~al.(2017)Kalchbrenner, Oord, Simonyan, Danihelka,
  Vinyals, Graves, and Kavukcuoglu]{kalchbrenner2017videopixel}
Nal Kalchbrenner, A{\"a}ron Oord, Karen Simonyan, Ivo Danihelka, Oriol Vinyals,
  Alex Graves, and Koray Kavukcuoglu.
\newblock Video pixel networks.
\newblock In \emph{Int. Conf. Mach. Learn.}, 2017.

\bibitem[Karras et~al.(2018)Karras, Aila, Laine, and Lehtinen]{pggan}
Tero Karras, Timo Aila, Samuli Laine, and Jaakko Lehtinen.
\newblock Progressive growing of gans for improved quality, stability, and
  variation.
\newblock In \emph{Int. Conf. Learn. Represent.}, 2018.

\bibitem[Karras et~al.(2019)Karras, Laine, and Aila]{stylegan}
Tero Karras, Samuli Laine, and Timo Aila.
\newblock A style-based generator architecture for generative adversarial
  networks.
\newblock In \emph{IEEE Conf. Comput. Vis. Pattern Recog.}, 2019.

\bibitem[Karras et~al.(2020{\natexlab{a}})Karras, Laine, Aittala, Hellsten,
  Lehtinen, and Aila]{karras2020analyzing}
Tero Karras, Samuli Laine, Miika Aittala, Janne Hellsten, Jaakko Lehtinen, and
  Timo Aila.
\newblock Analyzing and improving the image quality of stylegan.
\newblock In \emph{Proceedings of the IEEE/CVF conference on computer vision
  and pattern recognition}, pp.\  8110--8119, 2020{\natexlab{a}}.

\bibitem[Karras et~al.(2020{\natexlab{b}})Karras, Laine, Aittala, Hellsten,
  Lehtinen, and Aila]{stylegan2}
Tero Karras, Samuli Laine, Miika Aittala, Janne Hellsten, Jaakko Lehtinen, and
  Timo Aila.
\newblock Analyzing and improving the image quality of stylegan.
\newblock In \emph{IEEE Conf. Comput. Vis. Pattern Recog.}, pp.\  8110--8119,
  2020{\natexlab{b}}.

\bibitem[Karras et~al.(2021)Karras, Aittala, Laine, H{\"a}rk{\"o}nen, Hellsten,
  Lehtinen, and Aila]{karras2021alias}
Tero Karras, Miika Aittala, Samuli Laine, Erik H{\"a}rk{\"o}nen, Janne
  Hellsten, Jaakko Lehtinen, and Timo Aila.
\newblock Alias-free generative adversarial networks.
\newblock \emph{NIPS}, 34:\penalty0 852--863, 2021.

\bibitem[Kim et~al.(2020)Kim, Zhou, Philion, Torralba, and
  Fidler]{Kim2020_GameGan}
Seung~Wook Kim, Yuhao Zhou, Jonah Philion, Antonio Torralba, and Sanja Fidler.
\newblock {Learning to Simulate Dynamic Environments with GameGAN}.
\newblock In \emph{IEEE Conf. Comput. Vis. Pattern Recog.}, 2020.

\bibitem[Kim et~al.(2021)Kim, , Philion, Torralba, and
  Fidler]{Kim2021_DriveGAN}
Seung~Wook Kim, , Jonah Philion, Antonio Torralba, and Sanja Fidler.
\newblock {DriveGAN: Towards a Controllable High-Quality Neural Simulation}.
\newblock In \emph{IEEE Conf. Comput. Vis. Pattern Recog.}, 2021.

\bibitem[Kumar et~al.(2019)Kumar, Babaeizadeh, Erhan, Finn, Levine, Dinh, and
  Kingma]{kumar2019videoflow}
Manoj Kumar, Mohammad Babaeizadeh, Dumitru Erhan, Chelsea Finn, Sergey Levine,
  Laurent Dinh, and Durk Kingma.
\newblock Videoflow: A conditional flow-based model for stochastic video
  generation.
\newblock \emph{arXiv preprint arXiv:1903.01434}, 2019.

\bibitem[Lee et~al.(2018)Lee, Zhang, Ebert, Abbeel, Finn, and
  Levine]{lee2018stochastic}
Alex~X Lee, Richard Zhang, Frederik Ebert, Pieter Abbeel, Chelsea Finn, and
  Sergey Levine.
\newblock Stochastic adversarial video prediction.
\newblock \emph{arXiv preprint arXiv:1804.01523}, 2018.

\bibitem[Li et~al.(2022)Li, Wang, Snavely, and Kanazawa]{li2022infinitenature}
Zhengqi Li, Qianqian Wang, Noah Snavely, and Angjoo Kanazawa.
\newblock Infinitenature-zero: Learning perpetual view generation of natural
  scenes from single images.
\newblock \emph{arXiv preprint arXiv:2207.11148}, 2022.

\bibitem[Lin et~al.(2019)Lin, Gan, and Han]{lin2019tsm}
Ji~Lin, Chuang Gan, and Song Han.
\newblock Tsm: Temporal shift module for efficient video understanding.
\newblock In \emph{Int. Conf. Comput. Vis.}, pp.\  7083--7093, 2019.

\bibitem[Liu et~al.(2021)Liu, Tucker, Jampani, Makadia, Snavely, and
  Kanazawa]{liu2021infinite}
Andrew Liu, Richard Tucker, Varun Jampani, Ameesh Makadia, Noah Snavely, and
  Angjoo Kanazawa.
\newblock Infinite nature: Perpetual view generation of natural scenes from a
  single image.
\newblock In \emph{IEEE Conf. Comput. Vis. Pattern Recog.}, 2021.

\bibitem[Luc et~al.(2020)Luc, Clark, Dieleman, Casas, Doron, Cassirer, and
  Simonyan]{luc2020transformation}
Pauline Luc, Aidan Clark, Sander Dieleman, Diego de~Las Casas, Yotam Doron,
  Albin Cassirer, and Karen Simonyan.
\newblock Transformation-based adversarial video prediction on large-scale
  data.
\newblock \emph{arXiv preprint arXiv:2003.04035}, 2020.

\bibitem[Nash et~al.(2022)Nash, Carreira, Walker, Barr, Jaegle, Malinowski, and
  Battaglia]{nash2022transframer}
Charlie Nash, Jo{\~a}o Carreira, Jacob Walker, Iain Barr, Andrew Jaegle,
  Mateusz Malinowski, and Peter Battaglia.
\newblock Transframer: Arbitrary frame prediction with generative models.
\newblock \emph{arXiv preprint arXiv:2203.09494}, 2022.

\bibitem[Pan et~al.(2019)Pan, Wang, Jia, Shao, Sheng, Yan, and
  Wang]{pan2019video}
Junting Pan, Chengyu Wang, Xu~Jia, Jing Shao, Lu~Sheng, Junjie Yan, and
  Xiaogang Wang.
\newblock Video generation from single semantic label map.
\newblock In \emph{IEEE Conf. Comput. Vis. Pattern Recog.}, 2019.

\bibitem[Qiu et~al.(2022)Qiu, Jiang, Zhou, Wu, and Liu]{qiu2022stylefacev}
Haonan Qiu, Yuming Jiang, Hang Zhou, Wayne Wu, and Ziwei Liu.
\newblock Stylefacev: Face video generation via decomposing and recomposing
  pretrained stylegan3.
\newblock \emph{arXiv preprint arXiv:2208.07862}, 2022.

\bibitem[Radford et~al.(2016)Radford, Metz, and Chintala]{dcgan}
Alec Radford, Luke Metz, and Soumith Chintala.
\newblock Unsupervised representation learning with deep convolutional
  generative adversarial networks.
\newblock In \emph{Int. Conf. Learn. Represent.}, 2016.

\bibitem[Rakhimov et~al.(2020)Rakhimov, Volkhonskiy, Artemov, Zorin, and
  Burnaev]{rakhimov2020latent}
Ruslan Rakhimov, Denis Volkhonskiy, Alexey Artemov, Denis Zorin, and Evgeny
  Burnaev.
\newblock Latent video transformer.
\newblock \emph{arXiv preprint arXiv:2006.10704}, 2020.

\bibitem[Ren et~al.(2020)Ren, Chai, Tulyakov, Fang, Shen, and
  Yang]{ren2020human}
Jian Ren, Menglei Chai, Sergey Tulyakov, Chen Fang, Xiaohui Shen, and Jianchao
  Yang.
\newblock Human motion transfer from poses in the wild.
\newblock In \emph{Eur. Conf. Comput. Vis.}, 2020.

\bibitem[Ren \& Wang(2022)Ren and Wang]{ren2022look}
Xuanchi Ren and Xiaolong Wang.
\newblock Look outside the room: Synthesizing a consistent long-term 3d scene
  video from a single image.
\newblock In \emph{IEEE Conf. Comput. Vis. Pattern Recog.}, 2022.

\bibitem[Saito et~al.(2017)Saito, Matsumoto, and Saito]{saito2017tgan}
Masaki Saito, Eiichi Matsumoto, and Shunta Saito.
\newblock Temporal generative adversarial nets with singular value clipping.
\newblock In \emph{Int. Conf. Comput. Vis.}, 2017.

\bibitem[Saito et~al.(2020)Saito, Saito, Koyama, and Kobayashi]{saito2020train}
Masaki Saito, Shunta Saito, Masanori Koyama, and Sosuke Kobayashi.
\newblock Train sparsely, generate densely: Memory-efficient unsupervised
  training of high-resolution temporal gan.
\newblock \emph{Int. J. Comput. Vis.}, 2020.

\bibitem[Shen et~al.(2020)Shen, Yang, Tang, and Zhou]{shen2020interfacegan}
Yujun Shen, Ceyuan Yang, Xiaoou Tang, and Bolei Zhou.
\newblock Interfacegan: Interpreting the disentangled face representation
  learned by gans.
\newblock \emph{IEEE Trans. Pattern Anal. Mach. Intell.}, 2020.

\bibitem[Siarohin et~al.(2019{\natexlab{a}})Siarohin, Lathuilière, Tulyakov,
  Ricci, and Sebe]{Siarohin_2019_CVPR}
Aliaksandr Siarohin, Stéphane Lathuilière, Sergey Tulyakov, Elisa Ricci, and
  Nicu Sebe.
\newblock Animating arbitrary objects via deep motion transfer.
\newblock In \emph{IEEE Conf. Comput. Vis. Pattern Recog.}, 2019{\natexlab{a}}.

\bibitem[Siarohin et~al.(2019{\natexlab{b}})Siarohin, Lathuilière, Tulyakov,
  Ricci, and Sebe]{Siarohin_2019_NeurIPS}
Aliaksandr Siarohin, Stéphane Lathuilière, Sergey Tulyakov, Elisa Ricci, and
  Nicu Sebe.
\newblock First order motion model for image animation.
\newblock In \emph{Adv. Neural Inform. Process. Syst.}, 2019{\natexlab{b}}.

\bibitem[Skorokhodov et~al.(2022)Skorokhodov, Tulyakov, and
  Elhoseiny]{skorokhodov2022stylegan}
Ivan Skorokhodov, Sergey Tulyakov, and Mohamed Elhoseiny.
\newblock Stylegan-v: A continuous video generator with the price, image
  quality and perks of stylegan2.
\newblock In \emph{IEEE Conf. Comput. Vis. Pattern Recog.}, pp.\  3626--3636,
  2022.

\bibitem[Tian et~al.(2021)Tian, Ren, Chai, Olszewski, Peng, Metaxas, and
  Tulyakov]{tian2021mocoganhd}
Yu~Tian, Jian Ren, Menglei Chai, Kyle Olszewski, Xi~Peng, Dimitris~N Metaxas,
  and Sergey Tulyakov.
\newblock A good image generator is what you need for high-resolution video
  synthesis.
\newblock \emph{Int. Conf. Learn. Represent.}, 2021.

\bibitem[Tulyakov et~al.(2018)Tulyakov, Liu, Yang, and
  Kautz]{tulyakov2018mocogan}
Sergey Tulyakov, Ming-Yu Liu, Xiaodong Yang, and Jan Kautz.
\newblock Mocogan: Decomposing motion and content for video generation.
\newblock In \emph{IEEE Conf. Comput. Vis. Pattern Recog.}, 2018.

\bibitem[Unterthiner et~al.(2018)Unterthiner, van Steenkiste, Kurach, Marinier,
  Michalski, and Gelly]{unterthiner2018towards}
Thomas Unterthiner, Sjoerd van Steenkiste, Karol Kurach, Raphael Marinier,
  Marcin Michalski, and Sylvain Gelly.
\newblock Towards accurate generative models of video: A new metric \&
  challenges.
\newblock \emph{arXiv preprint arXiv:1812.01717}, 2018.

\bibitem[Van~den Oord et~al.(2016)Van~den Oord, Kalchbrenner, Espeholt,
  Vinyals, Graves, et~al.]{van2016pixelcnn}
Aaron Van~den Oord, Nal Kalchbrenner, Lasse Espeholt, Oriol Vinyals, Alex
  Graves, et~al.
\newblock Conditional image generation with pixelcnn decoders.
\newblock \emph{Adv. Neural Inform. Process. Syst.}, 2016.

\bibitem[Van~der Maaten \& Hinton(2008)Van~der Maaten and
  Hinton]{van2008visualizing}
Laurens Van~der Maaten and Geoffrey Hinton.
\newblock Visualizing data using t-sne.
\newblock \emph{Journal of machine learning research}, 9\penalty0 (11), 2008.

\bibitem[Vondrick et~al.(2016)Vondrick, Pirsiavash, and
  Torralba]{vondrick2016generating}
Carl Vondrick, Hamed Pirsiavash, and Antonio Torralba.
\newblock Generating videos with scene dynamics.
\newblock \emph{Adv. Neural Inform. Process. Syst.}, 2016.

\bibitem[Walker et~al.(2021)Walker, Razavi, and Oord]{walker2021prediction}
Jacob Walker, Ali Razavi, and A{\"a}ron van~den Oord.
\newblock Predicting video with vqvae.
\newblock \emph{arXiv preprint arXiv:2103.01950}, 2021.

\bibitem[Wang et~al.(2022{\natexlab{a}})Wang, Bau, and Zhu]{wang2022rewriting}
Sheng-Yu Wang, David Bau, and Jun-Yan Zhu.
\newblock Rewriting geometric rules of a gan.
\newblock \emph{ACM Transactions on Graphics (TOG)}, 41\penalty0 (4):\penalty0
  1--16, 2022{\natexlab{a}}.

\bibitem[Wang et~al.(2018)Wang, Liu, Zhu, Liu, Tao, Kautz, and
  Catanzaro]{wang2018video}
Ting-Chun Wang, Ming-Yu Liu, Jun-Yan Zhu, Guilin Liu, Andrew Tao, Jan Kautz,
  and Bryan Catanzaro.
\newblock Video-to-video synthesis.
\newblock \emph{arXiv preprint arXiv:1808.06601}, 2018.

\bibitem[Wang et~al.(2022{\natexlab{b}})Wang, Yang, Bremond, and
  Dantcheva]{wang2022latent}
Yaohui Wang, Di~Yang, Francois Bremond, and Antitza Dantcheva.
\newblock Latent image animator: Learning to animate images via latent space
  navigation.
\newblock \emph{arXiv preprint arXiv:2203.09043}, 2022{\natexlab{b}}.

\bibitem[Weissenborn et~al.(2019)Weissenborn, T{\"a}ckstr{\"o}m, and
  Uszkoreit]{weissenborn2019scaling}
Dirk Weissenborn, Oscar T{\"a}ckstr{\"o}m, and Jakob Uszkoreit.
\newblock Scaling autoregressive video models.
\newblock \emph{arXiv preprint arXiv:1906.02634}, 2019.

\bibitem[Wu et~al.(2021{\natexlab{a}})Wu, Huang, Zhang, Li, Ji, Yang, Sapiro,
  and Duan]{wu2021godiva}
Chenfei Wu, Lun Huang, Qianxi Zhang, Binyang Li, Lei Ji, Fan Yang, Guillermo
  Sapiro, and Nan Duan.
\newblock Godiva: Generating open-domain videos from natural descriptions.
\newblock \emph{arXiv preprint arXiv:2104.14806}, 2021{\natexlab{a}}.

\bibitem[Wu et~al.(2021{\natexlab{b}})Wu, Liang, Ji, Yang, Fang, Jiang, and
  Duan]{wu2021n}
Chenfei Wu, Jian Liang, Lei Ji, Fan Yang, Yuejian Fang, Daxin Jiang, and Nan
  Duan.
\newblock N$\backslash$" uwa: Visual synthesis pre-training for neural visual
  world creation.
\newblock \emph{arXiv preprint arXiv:2111.12417}, 2021{\natexlab{b}}.

\bibitem[Xiong et~al.(2018)Xiong, Luo, Ma, Liu, and Luo]{xiong2018learning}
Wei Xiong, Wenhan Luo, Lin Ma, Wei Liu, and Jiebo Luo.
\newblock Learning to generate time-lapse videos using multi-stage dynamic
  generative adversarial networks.
\newblock In \emph{Proceedings of the IEEE Conference on Computer Vision and
  Pattern Recognition}, pp.\  2364--2373, 2018.

\bibitem[Yan et~al.(2021)Yan, Zhang, Abbeel, and Srinivas]{yan2021videogpt}
Wilson Yan, Yunzhi Zhang, Pieter Abbeel, and Aravind Srinivas.
\newblock Videogpt: Video generation using vq-vae and transformers.
\newblock \emph{arXiv preprint arXiv:2104.10157}, 2021.

\bibitem[Yang et~al.(2018)Yang, Wang, Zhu, Huang, Shi, and Lin]{Yang_2018_ECCV}
Ceyuan Yang, Zhe Wang, Xinge Zhu, Chen Huang, Jianping Shi, and Dahua Lin.
\newblock Pose guided human video generation.
\newblock In \emph{Eur. Conf. Comput. Vis.}, 2018.

\bibitem[Yang et~al.(2021)Yang, Shen, and Zhou]{yang2021semantic}
Ceyuan Yang, Yujun Shen, and Bolei Zhou.
\newblock Semantic hierarchy emerges in deep generative representations for
  scene synthesis.
\newblock \emph{Int. J. Comput. Vis.}, 129\penalty0 (5):\penalty0 1451--1466,
  2021.

\bibitem[Yang et~al.(2022)Yang, Srivastava, and Mandt]{yang2022diffusion}
Ruihan Yang, Prakhar Srivastava, and Stephan Mandt.
\newblock Diffusion probabilistic modeling for video generation.
\newblock \emph{arXiv preprint arXiv:2203.09481}, 2022.

\bibitem[Yu et~al.(2022)Yu, Tack, Mo, Kim, Kim, Ha, and Shin]{yu2022dign}
Sihyun Yu, Jihoon Tack, Sangwoo Mo, Hyunsu Kim, Junho Kim, Jung-Woo Ha, and
  Jinwoo Shin.
\newblock Generating videos with dynamics-aware implicit generative adversarial
  networks.
\newblock \emph{arXiv preprint arXiv:2202.10571}, 2022.

\bibitem[Zhang et~al.(2022)Zhang, Peng, and Zhou]{zhang2022learning}
Qihang Zhang, Zhenghao Peng, and Bolei Zhou.
\newblock Learning to drive by watching youtube videos: Action-conditioned
  contrastive policy pretraining.
\newblock In \emph{European Conference on Computer Vision}, pp.\  111--128.
  Springer, 2022.

\end{thebibliography}
